\documentclass{gammarobot}
\usepackage[utf8]{inputenc}
\usepackage{xspace}
\usepackage{tabularx,array}
\usepackage{ragged2e}
\newcolumntype{L}[1]{>{\RaggedRight\arraybackslash}p{#1}}
\newcolumntype{Y}{>{\RaggedRight\arraybackslash}X}
\usepackage{enumitem}
\usepackage{fancyhdr}
\usetikzlibrary{arrows.meta,positioning,calc}

\setlist[itemize]{noitemsep,leftmargin=*}
\setlist[enumerate]{noitemsep,leftmargin=*}

\fancypagestyle{gammaTeleAI}{%
  \fancyhf{}
  \fancyhead[C]{\small\sffamily Gamma Robotics / TeleAI}
  \fancyfoot[C]{\thepage}
  \renewcommand{\headrulewidth}{\ifnum\value{page}>1 0.3pt\else 0pt\fi}
  
}
\makeatletter
\let\ps@plain\ps@gammaTeleAI
\makeatother

\makeatletter
\DeclareRobustCommand{\gscapegamma}{%
  \if b\expandafter\@car\f@series\@nil
    \ensuremath{\boldsymbol{\gamma}}%
  \else
    \ensuremath{\gamma}%
  \fi}
\makeatother
\newcommand{\gscape}{\texorpdfstring{\gscapegamma}{Gamma}-Scape\xspace}

\definecolor{gsink}{HTML}{263B4B}
\definecolor{gsblue}{HTML}{527C9C}
\definecolor{gsteal}{HTML}{477E78}
\definecolor{gsline}{HTML}{8799A6}

\title{\gscape: An Agent-Centric World for Mobile Embodied Intelligence}

\author[1,2,\ddagger]{Xiaotao Zhang}
\author[1,4,\ddagger]{Lin Qian}
\author[3,\ddagger]{Hao Zhang}
\author[1,2,\ddagger]{Xuanrui Cui}
\author[1]{Yuehao Huang}
\author[1]{Yunzi Wu}
\author[1]{Yucheng Deng}
\author[1]{Jiankun Dong}
\author[4]{Hujun Yin}
\author[1,\dagger]{Xinhai Li}
\author[1,5,*]{Chenjia Bai}
\author[1,*]{Xuelong Li}

\affiliation[1]{Institute of Artificial Intelligence (TeleAI), China Telecom}
\affiliation[2]{Shanghai Jiao Tong University}
\affiliation[3]{Carnegie Mellon University}
\affiliation[4]{The University of Manchester}
\affiliation[5]{Gamma Robotics}
\contribution{\textsuperscript{$\ddagger$}Equal Contributions\quad
  \textsuperscript{$\dagger$}Project Lead\quad
  \textsuperscript{*}Corresponding Authors}

\date{\today}
\metadata[Correspondence to]{Xiaotao Zhang (\email{dourakushinju@sjtu.edu.cn}); Chenjia Bai (\email{baicj@chinatelecom.cn})}

\abstract{
Recent embodied agents construct and adapt robot behavior through executable programs, tool use, and interaction feedback.
Studying these systems in mobile manipulation calls for physical worlds that connect agent-directed execution with its physical consequences and reusable interaction experience.
We introduce \gscape, an agent-centric simulation and benchmark platform for mobile embodied intelligence.
Agents compose motion tools and learned skills to carry out complete tasks that combine navigation and manipulation in a persistent physical world.
BenchForge grounds structured objectives in physical task instances with instructions, reference programs, and goal evaluators, defining task goals while preserving agents' choice of valid solutions.
An agentic collection harness further enables agents to produce demonstrations by carrying out contact-rich interactions through sustained control and physical feedback.
Assisted reference execution and human guidance provide complementary experience.
The resulting tool traces and aligned observation--action records support program refinement and policy learning, with common task-goal criteria for subsequent evaluation.
We benchmark representative agent systems on complete tasks and demonstrate contact collection and trace-informed program revision.
Together, these workflows provide infrastructure for research on embodied recursive self-improvement.
}

\begin{document}
\maketitle

\section{Introduction}
\label{sec:introduction}

Recent advances in embodied agents have broadened how robot behavior is constructed and adapted.
Agents generate executable programs over perception and control interfaces~\citep{codeaspolicies,capx}, use execution feedback to revise programs and retain reusable skills~\citep{aspire,rax}, and combine learned action skills with analytic motion tools~\citep{harnessvla}.
Mobile agents use multimodal observations and action interfaces to pursue goals across multiple locations~\citep{owmm}.

In mobile manipulation, each action changes the conditions for the next.
Opening a cabinet changes access, grasping occupies a hand, and transport changes the conditions for placement.
To retrieve supplies from storage and deliver them elsewhere, an agent must coordinate these interactions and respond to their measured outcomes.
Existing activity benchmarks and generative environments provide valuable tasks and demonstration resources~\citep{behavior1k,manitaskgen,robocasa365}.
We ask how a physical world can support complete-task evaluation while retaining the interaction context needed to construct and revise behavior.

\begin{figure}[t]
  \centering
  \includegraphics[width=\textwidth]{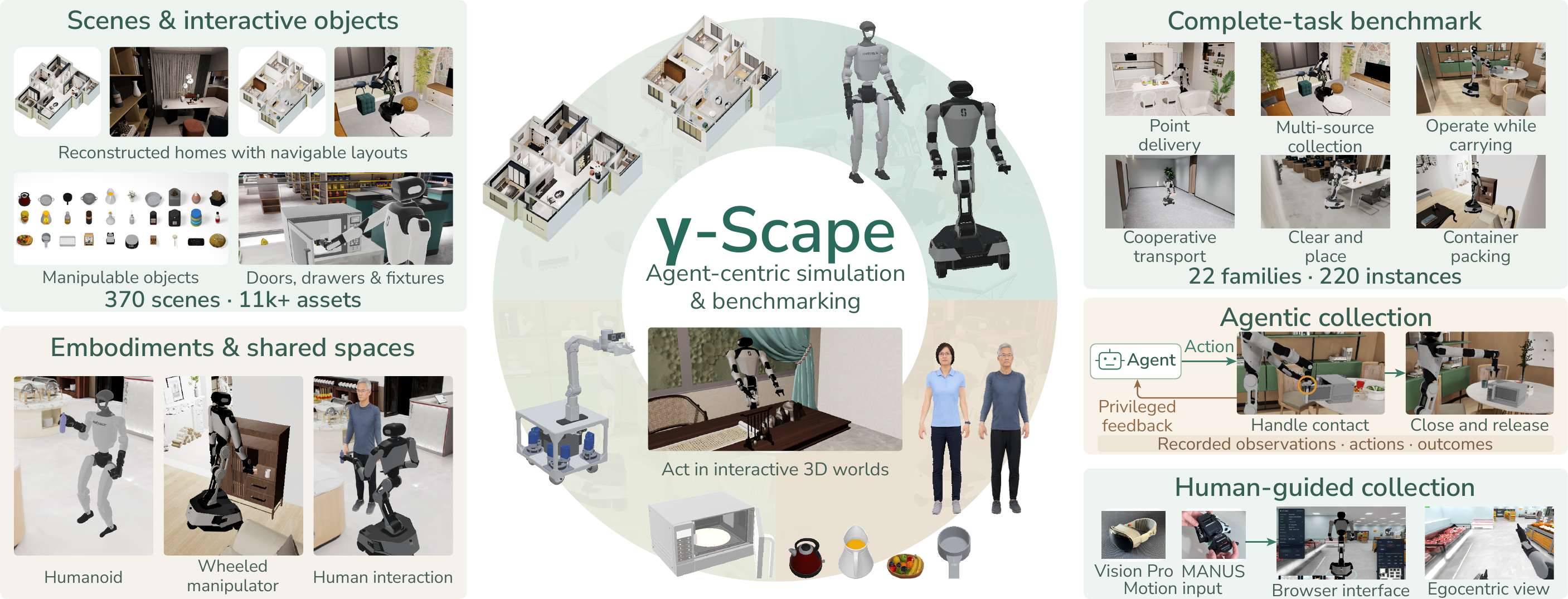}
  \caption{\textbf{An agent-centric world for mobile embodied intelligence.}
  \gscape brings diverse interactive worlds and embodiments together with complete-task benchmarking and experience collection.
  Privileged feedback guides agentic contact manipulation, while human teleoperation provides a complementary source of demonstrations.}
  \label{fig:overview}
\end{figure}

We introduce \textbf{\gscape}, an agent-centric simulation and benchmark platform for complete mobile manipulation tasks (\cref{fig:overview}).
The platform combines \textbf{370 indoor scenes} with \textbf{11,073 interactive object assets} for task construction.
Task agents compose their own programs and tool calls, combining motion tools and learned skills according to the states reached during execution.
\textbf{BenchForge} turns structured task requirements into physical instances by binding resources, constructing connected layouts, and applying geometric and physical checks.
Each instance includes an instruction, a reference program, and a goal evaluator.
The reference program supports construction and collection, while the evaluator checks requested outcomes and process conditions.
This separation lets researchers compare agent-chosen solutions under a fixed task definition and retain the same goals when revising behavior.
Different embodiments and shared-space participants provide additional physical contexts for studying these behaviors.

\gscape also supports agents as \textbf{producers of experience for behavior development}.
In this complementary workflow, native planning handles routine motion while a dedicated collection agent uses privileged physical feedback to control contact interactions.
It retains control through extraction, release, and withdrawal before task execution resumes from the measured state.
Each interaction yields both tool traces for program refinement and aligned robot observations and actions for skill adaptation.
Assisted reference execution and human guidance provide complementary demonstrations with explicit control-source and assistance labels.
The execution interfaces accept revised programs and learned skills for further runs under unchanged task goals, providing infrastructure for research on embodied recursive self-improvement.

We construct a fixed benchmark of \textbf{22 task families} and \textbf{220 task instances} spanning delivery, organization, storage and access, preparation and restoration, and carrying and coordination.
We compare representative agent systems as integrated into the platform, using common physical instances and task-goal criteria while retaining each system's reasoning and execution interfaces.
Complete-task success measures the requested outcome, while semantic progress records accumulated intermediate accomplishment.
Our experiments connect these results to execution traces and show how a failed contact interaction informs an executed program revision.
Our contributions are threefold:

\begin{itemize}
    \clubpenalty=10000
    \item \textbf{A world for agent-directed mobile manipulation.}
    We combine task-driven physical construction with composable execution interfaces, allowing agents to pursue explicit goals through programs, motion tools, and learned skills grounded in measured physical outcomes.
    \item \textbf{Experience production for behavior development.}
    We provide sustained agentic contact collection within complete tasks, alongside assisted and human-guided experience.
    We demonstrate contact collection and trace-informed program revision, retaining tool feedback and temporally aligned observation--action records for further adaptation.
    \item \textbf{A benchmark of complete embodied agent systems.}
    We construct a 22-family, 220-instance suite and examine representative systems through complete-task success, semantic progress, and physical execution analysis.
\end{itemize}

\section{Related Work}
\label{sec:related-work}

\paragraph{Embodied agents and behavior construction.}
Embodied agents use executable programs, tools, and interaction feedback to organize robot behavior.
Code as Policies~\citep{codeaspolicies} represents behavior through programs over perception and control interfaces, while CaP-X~\citep{capx} studies coding agents through interface abstractions and execution feedback.
ASPIRE~\citep{aspire} and RAX~\citep{rax} develop reusable behavior through program refinement and online exploration, respectively.
Harness VLA~\citep{harnessvla} combines a frozen vision-language-action model with analytic primitives and execution memory.
Mobile systems such as HomeRobot~\citep{homerobot}, BUMBLE~\citep{bumble}, and OWMM-Agent~\citep{owmm} integrate perception, reasoning, and action interfaces for tasks spanning multiple locations.
Vision-and-language navigation (VLN) studies instruction following in visual environments, spanning individual navigation and multi-robot coordination~\citep{navid,uninavid,spacevln,conavbench,deconav,cogddn,sparsevideonav,compass1,sagenav,navform,qwenrobotnav}.
For learned navigation, WNM-3D~\citep{wnm3d} conditions joint future-view and navigation-action generation on 3D scene representations.
\gscape complements these architectures with grounded tasks and a runtime for executing agent-chosen behavior under common goal criteria.

\paragraph{Interactive worlds and embodied benchmarks.}
ALFRED~\citep{alfred}, Habitat 2.0~\citep{habitat2}, and BEHAVIOR-1K~\citep{behavior1k} provide settings for instruction following, rearrangement, and everyday physical activities.
For navigation, GN0~\citep{gn0} connects data generation and policy learning with GN-Bench, a 3D Gaussian Splatting-based simulator and benchmark supporting closed-loop evaluation and dynamic human actors.
EMMOE~\citep{emmoe} focuses on mobile manipulation in open environments, while VLABench~\citep{vlabench} studies language-conditioned manipulation with long-horizon reasoning.
RoboCasa365~\citep{robocasa365} combines household mobile manipulation with demonstration data and policy evaluation, and LocoManiLab~\citep{locomanilab} brings together humanoid tasks, data, policies, controllers, and deployment.
We focus on complete-task behavior: agents select and compose operations while their physical effects persist across navigation, object handling, and fixture use.

\paragraph{Task generation and experience production.}
ManiTaskGen~\citep{manitaskgen} generates feasible mobile manipulation tasks for both evaluation and agent improvement.
RoboTwin~\citep{robotwin} and RoboTwin 2.0~\citep{robotwin2} connect simulated execution to demonstration synthesis; RoboTwin 2.0 further uses language models and simulation feedback to refine expert programs.
TEACh~\citep{teach} studies task-driven embodied interaction with a human partner, while EmbodiedCLBench~\citep{embodiedclbench} examines how accumulated experience transfers to new tasks.
\gscape couples task generation with agent-directed execution and dedicated contact collection in a shared physical runtime.
The resulting records preserve task context, tool feedback, and observation--action pairs for subsequent program and skill development.

\section{Method}
\label{sec:method}

\gscape connects task construction, execution, and experience production through a common task definition.
BenchForge grounds the physical instance and goal conditions; the runtime supports independent task-agent execution and reference-guided collection, retaining measured outcomes within the enclosing task.

\subsection{Simulation Environment}
\label{sec:simulation}

\gscape provides indoor environments that combine travel between interaction sites with object handling and fixture operation.
Scenes contain navigable floor space, furniture, movable objects, and articulated fixtures.
Meshes or Gaussian representations~\citep{gaussiansplatting} provide visual context; collision geometry, rigid bodies, and articulated assets enable physical interaction in an Isaac-based backend~\citep{isaacsim,isaaclab}.
A task can reuse suitable scene-native objects or insert selected assets, allowing its source, destination, and intervening interactions to be instantiated within one environment.

Interactive assets combine physical models with task-relevant spatial descriptions.
Movable objects support transport and rearrangement, while articulated fixtures provide constrained motion, including the rotation of cabinet doors and translation of drawers.
Fixture descriptions specify moving parts, joint types, axes, motion limits, handles, and associated support or receiving regions.
Opening a cabinet or pulling out a drawer changes access; the resulting object and joint states persist during navigation and manipulation.

Robot-mounted cameras provide RGB and, when configured, depth observations, while proprioception reports the robot's state.
The environment supports wheeled manipulators such as R1 Pro, humanoids such as G1, and quadruped-based manipulators.
Camera placement, body geometry, and reachable workspace determine each robot's available views and approaches, while body-specific controllers execute its motions.
Other robots and animated human actors share the scene, contributing both visible activity and physical occupancy in passages and work areas.

\begin{figure}[t]
  \centering
  \includegraphics[width=\textwidth]{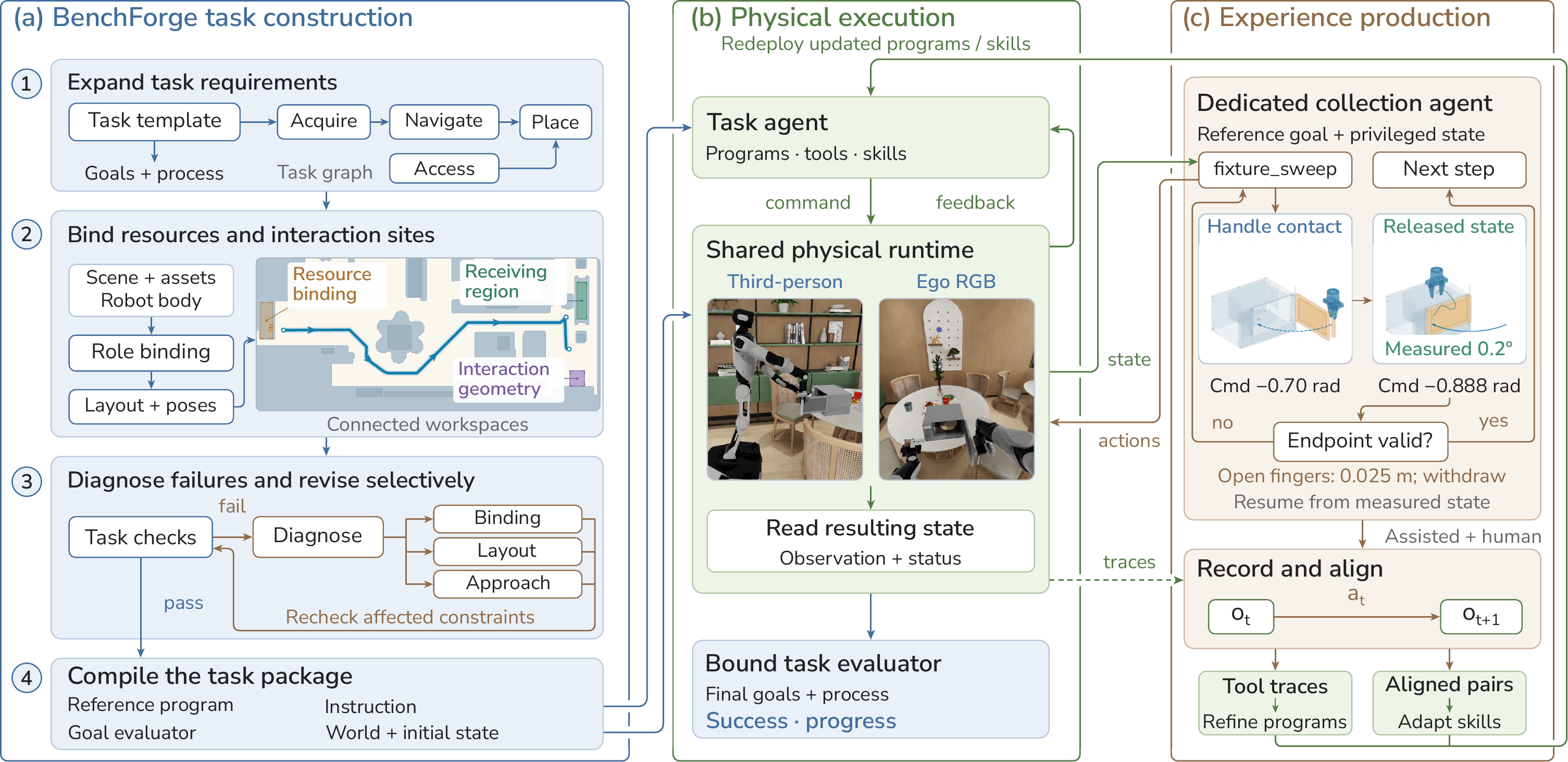}
  \caption{\textbf{From task objectives to reusable interaction experience.}
  (a) BenchForge expands task requirements, binds resources and interaction sites,
  and revises candidates using physical feedback before compiling the task package.
  (b) Task agents act through the shared physical runtime and receive measured state and execution feedback, illustrated by synchronized third-person and ego views;
  the bound evaluator assesses task success and semantic progress.
  (c) Dedicated collection exposes privileged context and disables manipulation assistance during agent-controlled contact.
  The contact views distinguish commanded hinge increments from the measured final angle.
  Execution resumes from the measured state, while aligned records support program and skill updates.
  Appendix~\ref{app:carrying-case} shows the complete unassisted carrying-and-articulation trial used in (b--c).
  The cross-room construction example in (a) is independent of that trial.}
  \label{fig:mechanisms}
\end{figure}

\subsection{Agent Execution}
\label{sec:execution}

For public task execution, agent-facing integrations expose observations, perception calls, motion tools, and learned skills while retaining each system's reasoning and memory mechanisms~\citep{codeaspolicies,capx}.
The agent organizes its own task solution through these interfaces.
A common runtime operates at robot-action boundaries while allowing agent-specific high-level tools.
A generated program can sequence and branch over tool calls; a learned skill can supply an action chunk within that program or act as a direct policy.
Motion tools, including those built on cuRobo~\citep{curobo}, translate spatial goals into trajectories, and embodiment adapters convert trajectory samples or policy outputs into the robot's control commands.
These execution forms share the loaded task and physical state.

Each command advances physics and updates task monitors from measured samples, possibly over multiple control steps.
Tools return observations, execution status, and terminal results; subsequent commands continue from the state actually reached.

The simulation clock also governs shared-space participants.
In step-driven execution, physics pauses during model inference and advances with actions; during real-time teleoperation, physics advances continuously.
A human actor's route, speed, and departure time determine its root transform at the current simulation time.
That transform drives both appearance and a kinematic collision volume; the actor remains at the route endpoint after arrival.
Task construction checks the actor's final occupancy against required access while permitting temporary obstruction along its route.
Multi-robot variants assign acquisition-to-release chains to individual robots, add their connecting navigation, and retain cross-robot prerequisites and shared-resource ordering.

\subsection{Task Generation}
\label{sec:task-generation}

\paragraph{Task specification and construction variables.}
BenchForge constructs tasks from an author-defined catalog of parameterized task-family templates (Fig.~\ref{fig:mechanisms}a).
Templates specify object and workspace roles, conditional phases and dependencies, allowed variants, and requested final-state and process conditions.
Instantiation activates the selected phases and expands repeated operations into a structured specification $\Gamma=(D,\mathcal R,\Phi)$, where $D=(V,E)$ is the operation graph, $\mathcal R$ contains object and workspace roles, and $\Phi$ specifies initial relations, final goals, and required processes.
Nodes in $D$ represent navigation, acquisition, placement, fixture actuation, or handover; edges encode prerequisites.
Given scene and asset resources $\mathcal E$ and robot embodiment $b$, a construction candidate is $\mathbf x=(\beta,\mathbf z,\mathbf q)$: $\beta$ binds roles to concrete resources, $\mathbf z$ contains initial layouts and reference object or fixture configurations, and $\mathbf q$ contains robot approaches and local interaction configurations.
For storage deposit, the roles identify an object, a source support, and a receiving compartment; opening and closing phases follow the requested fixture conditions.
Physical grounding resolves these configurations and the navigation connections between interaction sites.
The operation graph organizes construction and reference execution; the evaluated agent determines its own behavior.

\paragraph{Constraint-grounded instantiation.}
Let $\mathcal X(\Gamma,\mathcal E,b)$ be the candidate space and $\mathcal C_\Gamma$ the construction checks selected for the task and validation profile.
The problem is
\begin{equation}
\begin{aligned}
\text{find}\quad
&\mathbf x\in\mathcal X(\Gamma,\mathcal E,b)\\
\text{subject to}\quad
&c\!\left(\mathbf x_{\operatorname{scope}(c)}\right)=1,
\quad \forall c\in\mathcal C_\Gamma,
\end{aligned}
\label{eq:task-construction-problem}
\end{equation}
where $\operatorname{scope}(c)$ identifies the variable blocks used to instantiate check $c$.
The checks comprise resource, spatial, and physical conditions: $\mathcal C_\Gamma=\mathcal C_{\mathrm{res}}\cup\mathcal C_{\mathrm{spa}}\cup\mathcal C_{\mathrm{phy}}$.
Resource conditions match roles to supports, cavities, and mechanisms.
Spatial conditions enforce template-specific room assignments, workspace connectivity, and local capacity.
For example, with required workspaces $\mathcal W_D$, candidate base approaches $\mathcal A(\beta(w))$, and free space $\Omega_\rho$ adjusted by clearance $\rho$, site selection requires
\begin{equation}
\begin{aligned}
&\exists C\in\operatorname{CC}(\Omega_\rho)\ \text{such that}\\[-1pt]
&\mathcal A(\beta(w))\cap C\neq\varnothing,
\qquad\forall w\in\mathcal W_D.
\end{aligned}
\label{eq:construction-connectivity}
\end{equation}
Here $\operatorname{CC}$ denotes free-space connected components.
Capacity screening considers peak simultaneous occupancy along the reference operations; geometric instantiation then resolves placements and interaction configurations.
Physical trials assess settling, collisions, and the instantiated interactions selected for validation.

\paragraph{Diagnostic-guided refinement.}
BenchForge addresses Eq.~\ref{eq:task-construction-problem} through bounded candidate search: it selects resource bindings, instantiates local configurations, and revises candidates using geometric and physical diagnostics.
For a failed candidate $\mathbf x_k$, diagnosis $d_k$ selects editable variable blocks $\mathcal J(d_k)$.
An approach failure targets $\mathbf q$, a placement conflict targets the relevant part of $\mathbf z$, and a fixture--scene conflict can trigger rebinding in $\beta$.
The resulting repair neighborhood and affected checks are
\begin{equation}
\begin{aligned}
\mathcal N_{d_k}(\mathbf x_k)
&=\bigl\{\mathbf x_k\oplus\delta\ \bigm|\
\operatorname{supp}(\delta)\subseteq\mathcal J(d_k)\bigr\},\\
\mathcal C_{\mathrm{aff}}(\Delta_k)
&=\bigl\{c\in\mathcal C_\Gamma\ \bigm|\
\operatorname{scope}(c)\cap\Delta_k\neq\varnothing\bigr\}.
\end{aligned}
\label{eq:construction-refinement}
\end{equation}
Here $\operatorname{supp}(\delta)$ contains the blocks directly changed by edit $\delta$; $\oplus$ applies the edit and rebuilds dependent configurations.
For the selected edit $\delta_k$, $\Delta_k$ contains all blocks changed by $\mathbf x_k\mapsto\mathbf x_k\oplus\delta_k$, including regenerated configurations.
Checks are refreshed whenever they depend on a changed block.
These configuration dependencies can differ from the execution dependencies in $D$.
For example, a failed door check can identify an upstream storage placement for revision; changing that placement also refreshes the door check's configuration.
BenchForge searches the diagnosed repair neighborhood within the remaining budget and retains a candidate $\mathbf x^\dagger$ that passes the selected construction checks.

\paragraph{Task compilation and variation.}
\label{sec:task-families}
The accepted configuration determines the task package
\begin{equation}
\mathcal T_{\mathbf x^\dagger}
=\bigl(
\mathcal W_{\mathbf x^\dagger},\,
 s^0_{\mathbf x^\dagger},\,
 \ell_{\Gamma,\mathbf x^\dagger},\,
 P^{\mathrm{ref}}_{D,\mathbf x^\dagger},\,
 G_{\Phi,\mathbf x^\dagger}
\bigr),
\label{eq:compiled-task}
\end{equation}
comprising the physical world, initial state, instruction, reference program, and goal evaluator.
Language models assist with semantic and resource choices allowed by the template and refine the instruction.
The compiler derives all outputs from the instantiated specification and physical bindings.
The reference program follows $D$, whereas the evaluator binds the requested final and process conditions in $\Phi$ to actual entities and goal regions (Sec.~\ref{sec:evaluation}).
A reference landing pose provides a construction target within a receiving region; other placements satisfying the task remain valid.
Varying $\mathbf x$ under the same $\Gamma$ changes assets, layouts, and approaches while preserving task requirements.
Shared-space variants extend the construction with robot assignments and human routes that preserve required terminal access.
The fixed suite and task-family catalog are described in Sec.~\ref{sec:evaluation-setup} and Appendix~\ref{app:task-catalog}.

\subsection{Experience-Driven Adaptation}
\label{sec:learning}
\label{sec:collection}

Experience for program and skill development comes from public agent execution and dedicated collection workflows.
During agentic collection, the scheduler follows the task's reference structure through navigation and transport, then supplies a contact objective and privileged physical context to the collection agent (Fig.~\ref{fig:mechanisms}c).
With manipulation assistance disabled, the collection agent uses object and fixture geometry, receiving regions, and measured feedback to choose and revise robot-motion calls.
For example, an articulation-aware sweep uses the fixture's live joint geometry to guide wrist motion and returns the observed articulation and contact state.
Control remains with the agent through extraction or release and withdrawal, capturing the state left for subsequent task actions.
The scheduler resumes from the measured state once the operation's endpoint conditions hold; auxiliary interactions retain the parent objective.
Task success and collection completion are recorded separately; Appendix~\ref{app:collection-completion} specifies return conditions and deferred success termination.

Public execution supplies the observations and feedback exposed by each integration.
For program refinement, collection traces retain task context, ordered tool calls, arguments, returned feedback, and measured outcomes.
For skill adaptation, each executed action is paired with its pre-action observations; operation and segment boundaries preserve the scope of the recorded behavior.
Assisted execution by the reference executor (Oracle) and immersive human teleoperation contribute complementary demonstrations of complete tasks or focused interactions.
Control-source and assistance labels identify how each behavior was produced and guide experience selection.
Recording, temporal alignment, and export selection are specified in Appendix~\ref{app:learning}.

Integrated agent methods determine how execution feedback and tool traces inform program refinement.
Demonstration export and environment adapters connect learning systems such as LeRobot~\citep{lerobot} and RLinf~\citep{rlinfvla} to recorded observation--action samples and further interaction.
Revised programs run through the program interface, while learned policies use observation and action adapters to act directly or as skills invoked by a program.
Subsequent runs are assessed against unchanged task-goal criteria.
Sections~\ref{sec:challenge-analysis}--\ref{sec:platform-demonstrations} demonstrate contact collection and an executed program revision informed by retained experience.
These interfaces provide infrastructure for studying experience-driven embodied development and recursive self-improvement.

\section{Experiments}
\label{sec:experiments}
\label{sec:benchmark}

\subsection{Experimental Setup}
\label{sec:evaluation-setup}
\label{sec:evaluation}

The suite targets interaction-site selection, action prerequisites, and persistent physical conditions across navigation and manipulation.
Its 22 families form five groups: delivery and organization vary source--destination and spatial relations; storage and access add articulated mechanisms; preparation and restoration establish prerequisites or recover requested conditions; carrying and coordination constrain hand occupancy across actions (Appendix~\ref{app:task-catalog}).

\noindent
\begin{minipage}[t]{0.50\linewidth}
  \raggedright
  \vspace{0pt}
  The ten instances per family vary in objects, workspace bindings, layouts, and task variants.
  Scene sources include VLNVerse~\citep{vlnverse} and InteriorGS~\citep{interiorgs}.
  Table~\ref{tab:suite-profile} counts shared scenes once and includes background distractors in the asset count.
  We count scored semantic steps as operations, excluding navigation and technical transitions.
  For reference navigation distance, we sum the required route lengths in each compiled task.
\end{minipage}\hfill
\begin{minipage}[t]{0.46\linewidth}
  \vspace{0pt}
  \centering\small
\captionsetup{type=table,position=top,justification=raggedright}
\caption{\textbf{Coverage and task structure (220 instances).}
Rows 3 and 4 report the median and range.}
\label{tab:suite-profile}
\begin{tabular*}{\linewidth}{@{}l@{\extracolsep{\fill}}r@{}}
  \toprule
  \textbf{Statistic} & \textbf{Value}\\
  \midrule
  Distinct source scenes & 60\\
  Distinct object assets & 652\\
  Semantic operations per task & 3.5 / 1--8\\
  Reference navigation distance (m) & 8.07 / 1.11--61.01\\
  \bottomrule
\end{tabular*}

\end{minipage}

Each package fixes the instruction, initial state, and shared evaluator (Sec.~\ref{sec:task-generation}).
The reference Oracle completed all 220 instances end to end with manipulation assistance.
During construction, 70 instances (31.8\%) underwent candidate reselection or spatial revision (Appendices~\ref{app:task-catalog}--\ref{app:construction-example}).

We provide a reproducible evaluation package with simulator and BenchForge code, task templates, all 220 frozen task packages, evaluators, and baseline adapters.
Its suite index and setup instructions specify how to obtain and mount required scenes and assets from their source collections (Appendix~\ref{app:resources}).

The main comparison uses a single R1 Pro wheeled dual-arm robot with integrated versions of CaP-X~\citep{capx}, ASPIRE~\citep{aspire}, OWMM-Agent~\citep{owmm}, and RAX~\citep{rax}.
CaP-X, ASPIRE, and RAX use GPT-6-Astra; OWMM-Agent uses its released 8B checkpoint.
All systems use the same physical instances and evaluators, while retaining their own reasoning, perception, memory, and action interfaces.
Execution budgets use system-specific clocks and turn limits.
ASPIRE's skill library and RAX's exploration artifacts are developed outside the fixed suite and frozen before evaluation.
Agents receive instructions and their configured public observations and tools, excluding privileged collection context (Appendix~\ref{app:protocol}).
Virtual grasp and fixture attachments are disabled for all four systems.

Complete-task success requires the measured final state and execution history to satisfy the task's explicit conditions~\citep{behavior1k}.
For an executed trajectory $\tau$, let $\mathcal G$ contain final-state predicates and $\mathcal H$ contain required process predicates:
\begin{equation}
    \mathrm{Success}(\tau)=
    \left[\bigwedge_{g\in\mathcal G}g(\tau_{\mathrm{end}})\right]
    \land
    \left[\bigwedge_{h\in\mathcal H}h(\tau)\right].
    \label{eq:task-success}
\end{equation}
Each task binds its Boolean predicates to the named objects, receiving geometry, and requested outcomes.
Surface placement tests the horizontal bounds center and a lower-height bound, permitting intervening supports; region placement tests the full horizontal footprint, and cavity containment tests center-of-mass inclusion.
Secure-placement conditions require the designated spatial relation, release, and settling; $\tau_{\mathrm{end}}$ includes the specified observation window.
Process conditions, such as holding an object while opening a door, apply over their specified intervals, with no added requirement when $\mathcal H$ is empty.
The evaluator accepts any execution satisfying these conditions, including valid alternatives to the reference program (Appendix~\ref{app:criteria}).

Semantic progress summarizes achievement along the compiled task structure.
For positive endpoint weights $w_i$ and dependency-aware completion indicators $c_i\in\{0,1\}$,
\begin{equation}
    \mathrm{Progress}=\frac{\sum_i w_i c_i}{\sum_i w_i}.
    \label{eq:task-progress}
\end{equation}
Progress accumulates endpoint credit even if later actions reverse an intermediate result.
Endpoint weights are fixed at compilation and shared across systems; scored endpoints include navigation and manipulation, excluding technical transitions.
Valid alternatives may achieve task success with different reference-endpoint coverage; complete-task success remains the primary outcome.

In the standard environment, an episode ends automatically when the current task-success conditions are met.
An unsuccessful stop, irreversible violation, or exhausted budget ends the episode as a failure.
The final report records terminal outcomes and accumulated progress; infrastructure failures preventing evaluation are invalid.
Physics pauses during inference.
Invalid runs are recorded separately and replaced until each system--task pair has three valid trials.
Trial-produced histories, memory, and program revisions are reset between trials and tasks.
We report best-of-three, selecting one valid trial by success first and semantic progress second; both metrics use that trial.
Termination precedence, budget clocks, and system configurations are detailed in Appendix~\ref{app:protocol}.

\subsection{Benchmark Performance}
\label{sec:main-results}

Table~\ref{tab:main-results} reports overall performance and, for each task family, successful-task counts and summed semantic progress.
RAX achieves the highest overall success (21.8\%) and mean semantic progress (39.1\%), followed by ASPIRE (15.5\% / 34.5\%), CaP-X (10.0\% / 23.2\%), and OWMM-Agent (4.1\% / 12.2\%).
RAX completes at least one instance in 21 of the 22 task families, compared with 15 for ASPIRE, 11 for CaP-X, and 6 for OWMM-Agent.
ASPIRE completes one more instance than RAX in each of the spatial-arrangement and fixture-configuration families.
Summing the per-family maximum success counts gives a lower bound of 50 distinct instances completed without virtual bindings, spanning 21 families.

\begin{table}[htbp]
  \centering\small
  \caption{\textbf{Complete-task performance across the 22 task families.}
  Best of three valid trials per instance, ranked by success then progress;
  both metrics use the same trial, with invalid runs replaced.
  Family rows report success counts / summed progress over ten instances (0--10);
  final rows report overall success / mean progress (\%).
  Coordination denotes bimanual handling and inter-hand transfer by one robot.}
  \label{tab:main-results}
  \setlength{\tabcolsep}{6pt}\renewcommand{\arraystretch}{1.02}
  \begin{tabular*}{\linewidth}{@{\extracolsep{\fill}}llcccc@{}}
    \toprule
    \textbf{Group} & \textbf{Task family} & \textbf{CaP-X} & \textbf{ASPIRE} & \textbf{OWMM-Agent} & \textbf{RAX}\\
    \midrule
    \multirow{4}{*}{Delivery} & Point delivery & 3/4.50 & 4/6.50 & 2/3.00 & 4/5.50\\
    & Multi-source collection & 0/2.00 & 1/3.50 & 0/1.00 & 1/3.50\\
    & Multi-destination distribution & 1/2.75 & 1/3.75 & 0/1.00 & 2/4.50\\
    & Object exchange & 0/1.67 & 0/2.17 & 0/0.83 & 1/3.17\\
    \midrule
    \multirow{5}{*}{Organization} & Spatial arrangement & 2/3.00 & 4/5.50 & 1/1.50 & 3/4.00\\
    & Object sorting & 0/1.38 & 0/1.75 & 0/0.75 & 0/2.62\\
    & Stable stacking & 0/1.95 & 1/3.74 & 0/1.01 & 2/4.44\\
    & Container packing & 0/1.72 & 0/2.46 & 0/0.98 & 1/3.71\\
    & Clear and place & 1/3.00 & 1/3.75 & 0/1.25 & 2/4.75\\
    \midrule
    \multirow{4}{*}{Storage and access} & Fixture configuration & 5/5.00 & 6/6.00 & 2/2.00 & 5/5.00\\
    & Storage retrieval & 0/0.00 & 0/0.00 & 0/0.00 & 2/2.00\\
    & Storage deposit & 1/2.86 & 2/4.86 & 1/2.40 & 4/5.40\\
    & Storage transfer & 0/0.00 & 0/0.00 & 0/0.00 & 2/2.50\\
    \midrule
    \multirow{4}{*}{\shortstack[l]{Preparation and\\restoration}} & Route clearance & 4/5.05 & 6/7.52 & 2/3.00 & 6/6.52\\
    & Clear route and deliver & 1/3.23 & 1/3.77 & 0/1.49 & 1/4.21\\
    & Operation-space preparation & 2/3.94 & 3/5.30 & 1/2.61 & 4/5.64\\
    & Environment restoration & 0/1.77 & 0/2.55 & 0/0.98 & 1/3.95\\
    \midrule
    \multirow{5}{*}{\shortstack[l]{Carrying and\\coordination}} & Dual-object transport & 0/1.57 & 1/4.13 & 0/1.04 & 2/4.33\\
    & Cooperative large transport & 0/0.00 & 1/1.00 & 0/0.00 & 1/1.00\\
    & Handover and continue & 1/1.00 & 1/1.63 & 0/0.00 & 1/1.00\\
    & Operate while carrying & 1/2.59 & 1/3.49 & 0/0.89 & 2/4.30\\
    & Resource-constrained sequence & 0/2.02 & 0/2.61 & 0/1.02 & 1/3.97\\
    \midrule
    \multicolumn{2}{l}{\textbf{Overall success (\%)}} & 10.0 & 15.5 & 4.1 & 21.8\\
    \multicolumn{2}{l}{\textbf{Mean semantic progress (\%)}} & 23.2 & 34.5 & 12.2 & 39.1\\
    \bottomrule
  \end{tabular*}
\end{table}

Route-clearance success ranges from 20\% to 60\%, whereas clear-route-and-deliver success does not exceed 10\% for any system.
The carrying and coordination group has the lowest group-level success and mean semantic progress for all four systems; success rates are 4\% for CaP-X, 8\% for ASPIRE, 0\% for OWMM-Agent, and 14\% for RAX.
These results compare integrated systems, including their tools and budgets.
For example, OWMM-Agent's prompted vocabulary provides no explicit actions for fixture actuation or bimanual manipulation.

\begingroup\clubpenalty=10000
No system completes an object-sorting instance, although mean progress ranges from 7.5\% to 26.2\%.
\par\endgroup

Three recorded RAX episodes illustrate how local execution affects task continuation.
A RAX storage-retrieval trial resolves an IK failure with a higher torso-assisted approach and completes grasping, transport, and placement at 217.5 simulated seconds.
In storage deposit, the acquired puzzle is lost during a turn; recovery later ends in loss of wheel support.
In spatial arrangement, pose errors and locally feasible plans precede physical instability.
Both failed episodes terminate before their action budgets expire, with valid evaluator results (Appendix~\ref{app:baseline-case}).
Collision checking in RAX's free-space arm tool is restricted to robot self-collision.
The recorded cases show how planning feedback, contact outcomes, and base stability shape subsequent task execution across navigation, object transport, and final placement.

\begin{figure}[!b]
  \centering
  \includegraphics[width=\textwidth]{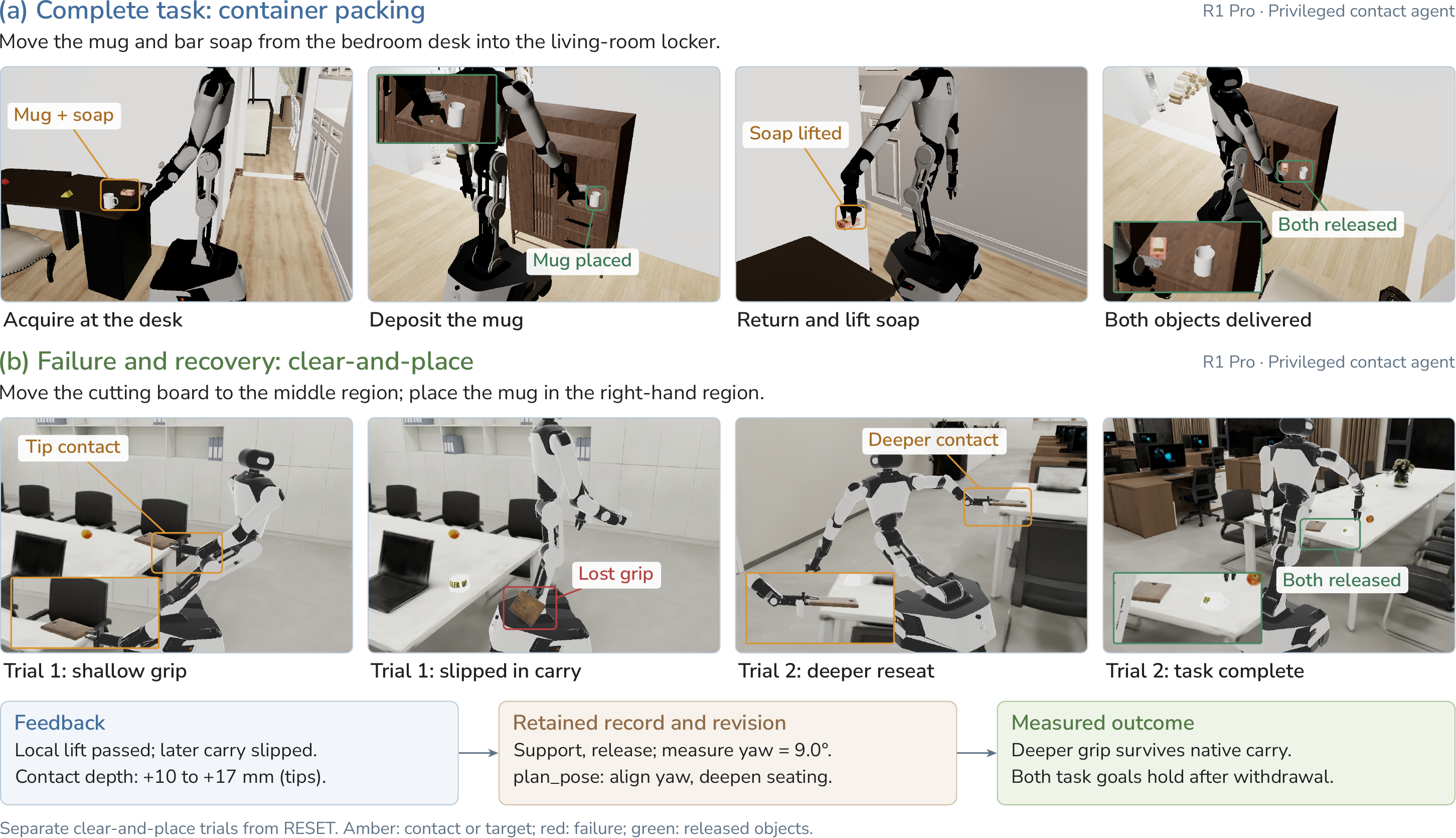}
  \caption{\textbf{From task execution to reusable experience.}
  (a) A completed container-packing trial connects acquisition at a bedroom
  desk with transport and placement in a living-room locker. Both objects remain
  in the locker after release and withdrawal.
  (b) A shallow board grasp slips during carrying. Its retained failure record
  informs a new trial after resetting the task, in which a deeper, yaw-aligned reseat enables
  successful task completion. The two attempts are separate rollouts.
  Both cases use a privileged contact agent and native motion planning without
  virtual grasp bindings. The panels are rendered from recorded simulation
  states.
  Appendix~\ref{app:program-revision} provides the recorded feedback and executed helper.}
  \label{fig:execution-experience}
\end{figure}

\subsection{Contact Execution and Task Continuation}
\label{sec:challenge-analysis}

The following two qualitative case studies examine contact collection using Codex CLI with GPT-6-Astra at high reasoning effort and the privileged context described in Appendix~\ref{app:collection}.
Native motion planning handles travel, while the collection agent controls grasping, lifting, placement, release, and withdrawal without virtual grasp bindings.
In container packing (Fig.~\ref{fig:execution-experience}a), R1 Pro carries a mug from a bedroom desk to a living-room locker, returns for bar soap, and deposits it beside the mug.
The initial mug-insertion plan is rejected; allowing the wrist orientation to vary yields an executable plan, and the subsequent placement leaves the mug settled upright on the shelf.
Both objects remain inside the locker after release and withdrawal, and the complete task passes its original evaluator.

\subsection{Experience Production and Reuse}
\label{sec:platform-demonstrations}

In clear-and-place (Fig.~\ref{fig:execution-experience}b), the initial board grasp passes a short lift but slips during native carry preparation, leaving the task unfinished.
An operator supplies the retained failure record to a new trial after resetting the task; the contact agent then selects and executes revised actions from new physical measurements.
When contact again shifts toward the fingertips, the agent supports and releases the board, then writes and executes a helper that aligns the approach to the measured board yaw and seats the edge deeper between the pads.
The run subsequently places both the board and mug in their designated regions, with task success maintained after release and withdrawal.

The original finger contacts lie near the pad tips, at wrist-frame depths of $[10,17]$\,mm.
After supported reseating, contacts span $[-34.8,17.3]$\,mm along the same axis, and the grasp survives both an approximately 42\,mm lift and native transport.

This case links a recorded contact failure to an executed program revision and subsequent task completion.
Appendix~\ref{app:program-revision} details the measured feedback and executed helper.

\begin{figure}[t]
  \centering
  \includegraphics[width=\linewidth]{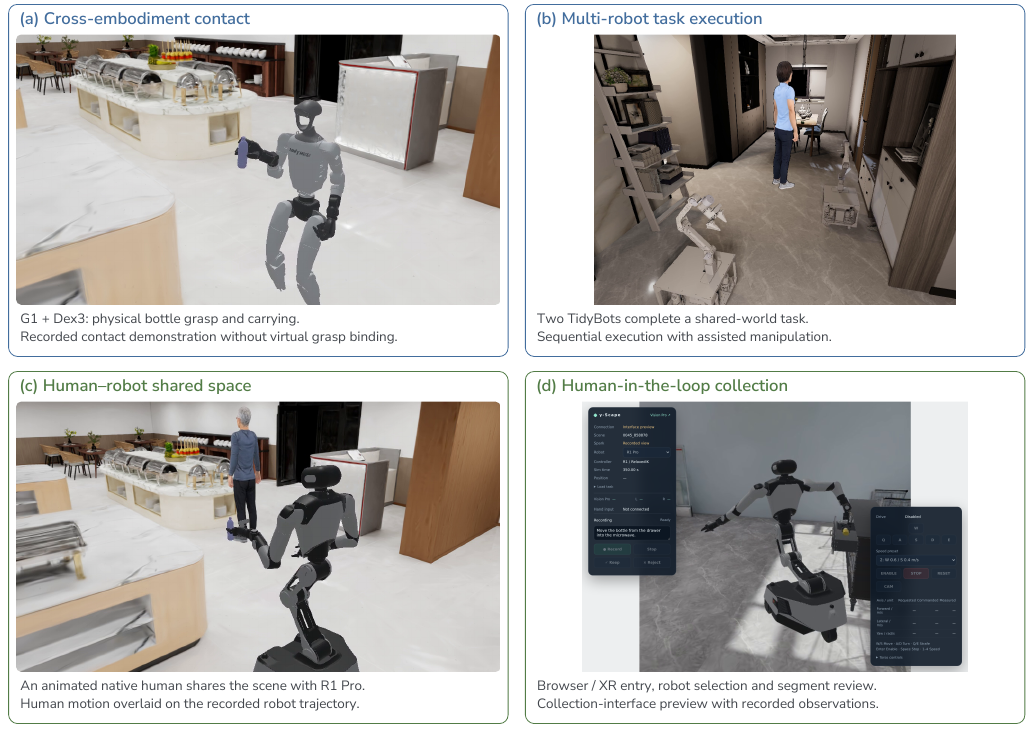}
  \caption{\textbf{Complementary platform settings.}
  (a) G1 physical bottle handling.
  (b) A two-TidyBot task with sequential assisted execution.
  (c) A native animated human alongside a recorded robot trajectory.
  (d) The collection interface displaying recorded observations.}
  \label{fig:platform-examples}
\end{figure}

Supplementary Fig.~\ref{fig:platform-examples} illustrates cross-embodiment contact, multi-robot execution, shared human--robot space, and the collection interface.

\section{Conclusion}
\label{sec:conclusion}

We introduced \gscape, an agent-centric simulation and benchmark platform for mobile embodied intelligence.
The platform connects grounded task construction, agent-directed execution, and reusable experience within persistent physical worlds.
BenchForge compiles task objectives into physical instances and explicit goal criteria.
Shared interfaces let agents compose programs, motion tools, and learned skills, with different solutions evaluated against the same task-goal criteria.

The benchmark results show that broad task-family coverage can coexist with low instance-level completion.
Carrying and coordination remain particularly challenging across the evaluated systems, while recorded episodes reveal how contact quality, reachability, and base stability affect subsequent actions.
Complete-task success and accumulated semantic progress distinguish satisfaction of task conditions from intermediate accomplishments, supporting analysis of both completed tasks and partial execution.

The collection studies connect local contact outcomes with subsequent transport and placement, and show how a retained failure record can inform an executable program revision.
Tool traces support program refinement, while aligned observation--action records from agentic, assisted, and human-guided execution provide material for skill adaptation.
Together, these capabilities connect behavior construction with experience reuse under common task goals.
They provide infrastructure for studying repeated adaptation and embodied recursive self-improvement.

\bibliographystyle{plainnat}
\bibliography{main}

\clearpage
\beginappendix

\section{Simulation Environment and Shared-Space Interaction}
\label{app:simulation}

\subsection{Scene Representation and Runtime}
\label{app:runtime}

Scene loaders support InteriorGS and VLNVerse visual assets together with occupancy maps and physical representations.
Task roles identify bodies, supports, cavities, links, joints, and handles.
Geometry queries use current object or link frames, so a drawer's receiving region follows its motion.
Embodiment adapters define camera placement, action channels, units, frames, rates, and command decoding.

The runtime uses Isaac Sim 5.1.0~\citep{isaacsim}, Isaac Lab 2.3.0~\citep{isaaclab}, and Python 3.11 on Linux.
R1 uses 120\,Hz physics and 30\,Hz control, with four physics substeps per action.
Goal monitors are updated from physical samples at each substep; navigation arrival and termination use the latest measurements at action boundaries.
Each observation includes its episode step and simulation time.
Cameras capture at 30\,Hz or on request; timestamps distinguish fresh from retained frames.
Explicit sensor refresh preserves the current simulation time and physical state.
Step-driven execution pauses physics between actions, including during model inference and rendering outside an action; during real-time teleoperation, physics advances continuously.

\paragraph{Resources and reproduction inputs.}
\label{app:resources}
We provide a reproducible evaluation package for the fixed 220-instance suite, including the simulator and BenchForge code, task templates, fixed-suite definitions, compiled task packages, evaluators, and baseline adapters.
Each fixed package contains its instruction, reset state and seed, reference program, and scoring conditions; the suite index maps task IDs to package locations.
The accompanying resource documentation and setup instructions identify the required scene and asset dependencies, explain how to obtain them from their source collections, and specify their resource mounts.
These dependencies follow the access and licensing terms of their source collections.
Interactive assets include resources from RoboCasa~\citep{robocasa} and SAPIEN's PartNet-Mobility collection~\citep{sapien,partnet,shapenet}.
With the resources mounted, evaluation loads the frozen packages to reproduce the specified instances.

\subsection{Human Motion and Physical Occupancy}
\label{app:human-motion}

A moving human actor is assigned a traversable polyline, speed, and departure time.
Let $\gamma:[0,L]\rightarrow\mathbb R^3$ parameterize the route by traveled distance.
For speed $v$ and departure time $t_0$, its position is
\begin{equation}
    \begin{aligned}
    \ell(t)&=\min\{L,\max\{0,v(t-t_0)\}\},\\
    \mathbf p(t)&=\gamma(\ell(t)).
    \end{aligned}
    \label{eq:human-motion}
\end{equation}
The actor faces along the active segment, waits before departure, and remains at the endpoint after arrival.
Walking animation advances with traveled distance; stationary activity clips advance with simulation time.
A common root transform updates rendering and an oriented kinematic collision box sized for the selected activity.

Route construction checks final occupancy against required task endpoints, operating footprints, and connectivity.
A conflicting endpoint is revised or excluded, while crossings and temporary obstruction remain permitted during execution.
Furniture-anchored activities retain their interaction location, so the surrounding task arrangement must accommodate the occupied workstation.

\subsection{Multi-Robot Task Structure}

An acquisition-to-release handling chain forms one assignment unit; standalone fixture operations can be assigned separately.
Task construction assigns these units to robots, preserves object-handling dependencies, and specifies each robot's connecting navigation, initial state, and parking state.
Reference scheduling enforces $t_j^{\mathrm{start}}\geq t_i^{\mathrm{end}}$ when operation $j$ depends on $i$.
Shared-resource ordering and conflicts between reference navigation curves can introduce waits at operation boundaries.
Reference timing sets the operation order, while body-specific controllers execute the motions.
Multi-robot variants use distinct robots; the fixed benchmark's two-hand tasks use one robot.

\section{Agentic Contact Collection}
\label{app:interfaces}
\label{app:collection}

\subsection{Contact-Collection Interface}

\paragraph{Collection agent.}
The collection cases in Secs.~\ref{sec:challenge-analysis}--\ref{sec:platform-demonstrations} use the system-installed Codex CLI with GPT-6-Astra at high reasoning effort, independently of the four benchmark baselines.
The harness starts one Codex process with the task instruction, compiled contact objectives, robot description, and tool definitions.
Codex receives the privileged context in Table~\ref{tab:contact-context}, inspects images and measured contacts, writes numerical helpers, and invokes the existing motion and gripper tools.
Native scheduling handles travel between interactions; Codex retains one conversation across the task's contact operations.
These trials allow up to 200 episode-tool requests and impose no additional collection wall-time limit; physics pauses during reasoning.
An operator launches each trial and supplies its initial brief.
Within the rollout, Codex selects and executes the contact adjustments without manual robot control.
A new RESET starts a new agent workspace and conversation; prior experience is supplied explicitly, as in Sec.~\ref{app:program-revision}.
The initial prompt, tool requests and replies, generated helpers, action records, and camera streams are retained.

Once a contact operation is ready, the scheduler provides it to the collection agent together with its parent context and the information in Table~\ref{tab:contact-context}.
With manipulation assistance disabled, the agent invokes robot-motion tools for approach, gripper actuation, and local adjustment.
Native execution handles the surrounding navigation and transport.
Subsequent observations report the resulting object and fixture states.
For each executed command, the runtime checks that the returned robot snapshot corresponds to the final simulation step.
The pose tool separately reports planned samples, executed actions, remaining motion, and the observed state.
Before the next motion request, the collection tool refreshes planning geometry and held-object geometry from current measurements.

\begin{table}[htbp]
  \centering\small
  \caption{\textbf{Contact-collection context and its use.}}
  \label{tab:contact-context}
  \renewcommand{\arraystretch}{1.12}
  \begin{tabularx}{\linewidth}{@{}L{.28\linewidth}Y@{}}
    \toprule
    Context & Use in the interaction\\
    \midrule
    Active objective & Identify the target, requested relation or articulation, and parent operation.\\
    Robot state & Select a hand, approach pose, base adjustment, and feasible joint motion.\\
    Object geometry & Inspect pose, extent, nearby collision geometry, and physical properties.\\
    Fixture geometry & Locate the handle, joint axis, pivot, limits, and current articulation.\\
    Receiving geometry & Choose a pose within the permitted support, region, or cavity.\\
    Measured feedback & Inspect images, contacts, object motion, and the requested physical effect.\\
    \bottomrule
  \end{tabularx}
\end{table}

Assisted reference execution follows generated programs using interaction assistance, including virtual grasp bindings.
Control-source and assistance labels distinguish these records from contact-driven experience during export (Sec.~\ref{app:learning}).

\subsection{Operation Lifecycle and Session Completion}
\label{app:collection-completion}

For acquisition, a contact operation continues through lifting the object and leaving the source; for placement, it continues through release and withdrawal.
Fixture interaction covers the requested articulation, and transfer covers the corresponding multi-hand goal.
The agent can request supporting operations while the runtime retains the parent objective and its monitoring.
When the scheduler accepts the return from a supporting operation, it resumes the parent operation; task credit remains unchanged.
It advances beyond the top-level operation only after the original physical endpoint predicates are satisfied, continuing from the measured state.

Recordings distinguish operation return, task success, and collection completion.
The evaluator determines task success, while collection completion indicates that all scheduled operations have returned and all child operations are inactive.

Contact collection defers automatic success termination while retaining the original goal predicates, failure conditions, and simulation-time limit.
This allows release and withdrawal to continue after a goal first becomes true.
Once scheduled and nested operations have finished, successful collection requires the current task goal to satisfy the configured consecutive-success count, one action by default.
Further physics steps complete any pending stability windows; if the schedule ends with unsatisfied goals, the task is recorded as unsuccessful.
A manual finish can end collection before the schedule is complete, with the current task result and collection-completion status recorded separately.

\subsection{Articulation-Aware Motion Tools}

For a revolute joint, let $\mathbf c$ lie on its world-space axis, $\mathbf a$ be the unit axis direction, and $(\mathbf p_0,R_0)$ be the measured wrist pose when a sweep is requested.
An angular increment $\Delta\theta$ defines the wrist target
\begin{equation}
    \begin{aligned}
    \mathbf p^\star&=\mathbf c+R_{\mathbf a}(\Delta\theta)(\mathbf p_0-\mathbf c),\\
    R^\star&=R_{\mathbf a}(\Delta\theta)R_0,
    \end{aligned}
    \label{eq:fixture-sweep}
\end{equation}
where $R_{\mathbf a}$ rotates about the fixture axis.
For a prismatic joint, the target is $\mathbf p^\star=\mathbf p_0+\Delta d\,\mathbf a$, with unchanged wrist orientation.
The planner converts these targets into robot motions subject to kinematic and collision constraints, including permitted base and torso motion.
The tool returns the measured joint and contact state after execution.
The geometric target guides robot motion, and contact between the robot and fixture produces the joint displacement.

\section{Task Construction and Goal Semantics}
\label{app:construction}

\subsection{Template Expansion and Resource Binding}

Template parameters control object multiplicity, spatial scope, fixture configuration, and whether the task requires restoring the environment after intermediate changes.
Instantiation activates conditional phases, expands repeated operations, and establishes their prerequisites in the operation graph $D$.
Physical grounding binds object and workspace roles to usable scene-native objects or inserted assets through their support domains, cavity domains, and fixture joint and handle geometry.
It also supplies navigation connections between interaction locations.
Language models assist with template-permitted choices, such as selecting compatible assets or choosing sorting categories supported by the available objects, and refine the corresponding instruction.
The selected semantic options and physical bindings determine the shared specification from which the compiler derives the instruction, reference operations, and requested goal and process conditions.

\paragraph{Room grouping.}
Let $\mathcal W_D$ contain the workspace roles required by the operation graph $D$, and let $\beta(w)$ bind workspace $w$ to a compatible resource.
For cross-room templates, a group label $\eta(w)$ specifies which workspaces should share a room.
The assignments satisfy
\begin{equation}
\begin{aligned}
&\operatorname{room}(\beta(w_i))=\operatorname{room}(\beta(w_j))\\[-1pt]
&\qquad\Longleftrightarrow\quad \eta(w_i)=\eta(w_j),\quad
w_i,w_j\in\mathcal W_D.
\end{aligned}
\label{eq:workspace-room-groups}
\end{equation}
Thus, related workspaces remain co-located while different groups span distinct rooms.
Local and cross-workspace templates retain their respective spatial requirements.

\subsection{Connected Layout Search and Failure-Directed Repair}

Connected workspace assignments undergo local capacity screening before detailed geometric checks and configured physical trials.

\paragraph{Workspace capacity.}
Tabletop capacity is checked against the objects that must occupy a region simultaneously.
Let $\mathcal O_g^{(k)}$ contain the objects assigned to a named region $g$ after reference operation $k$, with $k=0$ denoting the initial layout and $K$ the number of reference operations.
For support $s$ of size $(W_s,H_s)$, let $\mathcal Q_{g,s}(n)$ contain padded region-seed proposals for $n$ occupants whose dimensions fit that support within tolerance $\epsilon$.
The capacity filter is
\begin{equation}
\begin{aligned}
n_g &= \max_{0\leq k\leq K}\bigl|\mathcal O_g^{(k)}\bigr|,\\
a_{g,s} &= \min_{(u,v)\in\mathcal Q_{g,s}(n_g)} uv,\\
\sum_{g\mapsto s}a_{g,s} &\le (W_s+\epsilon)(H_s+\epsilon),
\end{aligned}
\label{eq:workspace-capacity}
\end{equation}
where $g\mapsto s$ denotes assignment to the same physical support. A binding is rejected if its proposal set is empty.
Peak occupancy permits sequential reuse of a surface, while the summed seed areas screen out overcommitted supports before detailed placement and interaction checks.
The dimension tolerance is $\epsilon=0.01$\,m.
This filter applies to named tabletop regions with known object dimensions; floor workspaces bypass the capacity check.

\paragraph{Failure-directed repair.}
Failures trigger revisions to the responsible variables within the construction budget (Table~\ref{tab:construction-repair}).
Geometric compilation, physical endpoint trials, and complete reference executions are recorded as distinct validation stages.
Shared-space construction follows the route and assignment procedures in Sec.~\ref{app:simulation}.

\begin{table}[htbp]
  \centering\small
  \caption{\textbf{Construction feedback and revision scope.} Repairs preserve the requested task relations while changing the relevant physical variables.}
  \label{tab:construction-repair}
  \renewcommand{\arraystretch}{1.12}
  \begin{tabularx}{\linewidth}{@{}L{.36\linewidth}Y@{}}
    \toprule
    Failed condition & Candidate revision\\
    \midrule
    Local placement or overlap & Resample the object pose and refresh dependent endpoints.\\
    Approach or interaction geometry & Change stance, approach side, or interaction configuration.\\
    Support, cavity, or asset compatibility & Rebind the resource or select a compatible asset.\\
    Site access or connectivity & Select alternative connected sites or navigation connections.\\
    Physical endpoint execution & Use the diagnostic to revise its pose, resource, or site.\\
    Terminal human occupancy & Revise the route endpoint to preserve required access.\\
    \bottomrule
  \end{tabularx}
\end{table}

\subsection{Task Families and Fixed Suite}
\label{app:task-catalog}

Table~\ref{tab:task-catalog} explains the five task groups and defines their 22 families.
The groups combine shared physical operations under different spatial, access, and manipulation-resource constraints.
The fixed suite contains ten instances per family; object choices, workspace bindings, layouts, and permitted variants provide within-family variation.
A source scene denotes the underlying indoor environment and may support multiple task instances with different selected objects, initial configurations, destinations, and requested outcomes.
Two-hand possession, handover, and restoration are evaluated only when requested by the task.

Table~\ref{tab:construction-statistics} summarizes the retained source-generation attempts for the 220 fixed-suite tasks.
Candidate reselection counts instances with at least one rejected candidate for a task-construction reason; service retries are excluded from this count.
Spatial revision counts instances whose selected candidate required more than one layout pass, including changes to stance, interaction side, workspace, or scene.
The counts overlap in 11 instances, yielding 70 distinct packages with either revision (31.8\%).
The latency timing window spans the retained generation attempt through package finalization, including model calls, initialization, configured physical checks, and packaging.
The reported duration excludes recorded simulator-slot waits; full reference-Oracle execution and earlier discarded outer generation attempts fall outside this timing window.

\begin{table}[htbp]
  \centering\small
  \caption{\textbf{Construction records for retained fixed-suite packages.}}
  \label{tab:construction-statistics}
  \renewcommand{\arraystretch}{1.10}
  \begin{tabular*}{\linewidth}{@{}l@{\extracolsep{\fill}}r@{}}
    \toprule
    \multicolumn{2}{@{}l@{}}{\textit{Automatic revisions ($n=220$)}}\\
    \addlinespace[2pt]
    Candidate reselection & 25 / 220 (11.4\%)\\
    Spatial layout revision & 56 / 220 (25.5\%)\\
    Either revision & 70 / 220 (31.8\%)\\
    \midrule
    \multicolumn{2}{@{}l@{}}{\textit{Construction latency ($n=220$)}}\\
    \addlinespace[2pt]
    Median & 4.94 min\\
    Mean & 5.81 min\\
    90th percentile & 7.34 min\\
    \bottomrule
  \end{tabular*}
\end{table}

\begin{table}[htbp]
  \centering\small
  \caption{\textbf{Task groups and families.} Cooperative large transport
  uses two hands of one robot; multi-robot variants are a separate extension.}
  \label{tab:task-catalog}
  \renewcommand{\arraystretch}{1.10}
  \begin{tabularx}{\linewidth}{@{}>{\raggedright\arraybackslash}p{.27\linewidth}>{\raggedright\arraybackslash}X@{}}
    \toprule
    \textbf{Family} & \textbf{Objective}\\
    \midrule
    \multicolumn{2}{@{}>{\raggedright\arraybackslash}p{\linewidth}@{}}{\textbf{Delivery.} Links acquisition, transport, and release across sources and destinations. Multiple sources, destinations, or exchanged objects require organizing visit order and delivery under specified object--destination assignments.}\\
    \addlinespace[2pt]
    Point delivery & Move one object from its source to a destination and release it stably.\\
    Multi-source collection & Collect specified objects from multiple sources at a common destination.\\
    Multi-destination distribution & Deliver objects from a common source to their assigned destinations.\\
    Object exchange & Exchange two objects' source positions, allowing valid temporary staging.\\
    \midrule
    \multicolumn{2}{@{}>{\raggedright\arraybackslash}p{\linewidth}@{}}{\textbf{Organization.} Establishes requested spatial, grouping, support, or containment relations. Individual placements contribute to a consistent final arrangement, and earlier placements can affect subsequent operations.}\\
    \addlinespace[2pt]
    Spatial arrangement & Establish specified object regions or spatial relations.\\
    Object sorting & Group objects by category on the receiving surface.\\
    Stable stacking & Establish the required object-support relations in a stable arrangement.\\
    Container packing & Put all specified objects inside an open container's cavity.\\
    Clear and place & Relocate an occupying object and place the target at the cleared destination.\\
    \midrule
    \multicolumn{2}{@{}>{\raggedright\arraybackslash}p{\linewidth}@{}}{\textbf{Storage and access.} Connects fixture states with object handling. Agents establish access before retrieval or deposit when needed and restore the fixture when requested.}\\
    \addlinespace[2pt]
    Fixture configuration & Move a specified furniture joint to its requested state.\\
    Storage retrieval & Obtain an object from storage and deliver it, opening a door or drawer when needed.\\
    Storage deposit & Place an object inside a storage compartment; close operated joints when requested.\\
    Storage transfer & Move an object between storage compartments, restoring operated joints when requested.\\
    \midrule
    \multicolumn{2}{@{}>{\raggedright\arraybackslash}p{\linewidth}@{}}{\textbf{Preparation and restoration.} Modifies the environment to enable another action, such as clearing a route or preparing an operation area. Restoration requires recovering specified configurations after intermediate changes.}\\
    \addlinespace[2pt]
    Route clearance & Relocate an obstacle so the required route becomes usable.\\
    Clear route and deliver & Clear the route and deliver a payload, restoring displaced objects when requested.\\
    Operation-space preparation & Relocate an object that blocks a fixture operation, then complete the operation.\\
    Environment restoration & Return designated objects and fixtures to explicit target configurations.\\
    \midrule
    \multicolumn{2}{@{}>{\raggedright\arraybackslash}p{\linewidth}@{}}{\textbf{Carrying and coordination.} Introduces occupied-hand constraints, bilateral manipulation, possession transfer, and operations performed while retaining another object. Agents coordinate these actions under limited manipulation resources.}\\
    \addlinespace[2pt]
    Dual-object transport & Hold two objects in separate hands with required overlap and complete delivery.\\
    Cooperative large transport & Use both hands to support, transport, and place one large object.\\
    Handover and continue & Transfer an object between hands and complete its receiving-hand placement goal.\\
    Operate while carrying & Maintain a payload in one hand while operating a fixture with the other, then deliver it.\\
    Resource-constrained sequence & Complete delivery and an intermediate operation under hand-capacity requirements.\\
    \bottomrule
  \end{tabularx}
\end{table}

\subsection{Recorded Construction Revision}
\label{app:construction-example}

\paragraph{Interaction-side revision.}
The retained construction record for \texttt{container\_packing\_7008107074616442} in InteriorGS \texttt{0642\_841452} begins with a rejected layout.
Joint stance search and a forward-shifted interaction configuration also fail to compile the placement endpoints.
Changing the interaction side while retaining the scene and selected tabletop produces a compiled layout; its configured physical validation then passes on the first attempt.
The revision changes how the objects and robot are arranged for interaction, while preserving the requested packing operations.

\paragraph{Shared-space access.}
A shared-space construction example uses a two-robot collection task in VLNVerse scene \texttt{kujiale\_0087}.
The original human route ended at $(3.08435,3.61516)$\,m, obstructing a required navigation connection when the stationary teammate was included.
The export pass revised the human route and its terminal position to $(2.93435,5.76516)$\,m while retaining the human's starting position, speed of 0.6\,m/s, and the task's operation ordering.
Connectivity was rechecked with the recorded route clearance and the teammate's occupancy before recompilation and reference execution.

\begin{table}[htbp]
  \centering\small
  \caption{\textbf{A recorded construction revision.} One additional shared-space task; coordinates share the recorded world frame.}
  \label{tab:construction-example}
  \begin{tabularx}{\linewidth}{@{}L{.30\linewidth}Y@{}}
    \toprule
    Stage & Saved observation or outcome\\
    \midrule
    Initial candidate & Human terminal occupancy and the stationary teammate disconnect a required route.\\
    Editable variables & Human route and endpoint; starting position, speed, task goals, and operation order unchanged.\\
    Revised endpoint & $(2.93435,5.76516)$\,m; revised connection length 4.786\,m.\\
    Full reference run & RESET to task success; 246.02 simulated seconds; assisted, sequential two-robot execution.\\
    \bottomrule
  \end{tabularx}
\end{table}

\subsection{Final-State and Process Conditions}
\label{app:criteria}

Each compiled task binds a conjunction of conditions to its named entities and requested outcomes.
Final-state predicates use current measurements or a terminal stability window; process predicates retain required events and invariants over their specified intervals.
The following examples map instruction excerpts from five existing task packages to their actual semantic bindings.
Table~\ref{tab:goal-semantics} defines the component measurements, geometric rules, and default thresholds.

Object identities remain fixed in the four object-handling examples.
Hand labels in possession conditions identify the effector bindings stored in each task package; placement release requires that neither hand holds the specified object.
``Settled'' denotes the task's 0.5\,s post-release window; geometric relations follow Table~\ref{tab:goal-semantics}.

\begin{description}
  \setlength{\itemsep}{5pt}
  \item[Storage deposit.] \emph{``Place the stapler securely inside the drawer, then close it.''}
    Stapler COM inside the named moving cavity; object release and settling conditions satisfied; drawer closed; destination-workspace membership.
  \item[Point delivery.] \emph{``\ldots set the toy bus down securely on it.''}
    Toy bus \texttt{on\_surface} of the destination table; object release and settling conditions satisfied; destination-workspace membership.
  \item[Stable stacking.] \emph{``\ldots approach the cutting board, and set the mug down securely on it.''}
    Board \texttt{on\_surface} of the destination table and mug \texttt{on\_object} of the live board; object release and settling conditions satisfied for both objects; both in the destination workspace. Both object outcomes remain required at completion.
  \item[Fixture configuration.] \emph{``\ldots then open it and leave it open.''}
    The specified top drawer has the semantic open state; this is the task's only condition.
  \item[Operate while carrying.] \emph{``While holding the cube in your left hand, use your other hand to open the door, leaving it open.''}
    \emph{During opening:} left-hand possession throughout the monitored door-opening operation. \emph{At completion:} cube \texttt{in\_region} of the specified right-side tabletop region; object release and settling conditions satisfied; cabinet door open; destination-workspace membership.
\end{description}

\begin{table}[!htb]
  \centering\small
  \caption{\textbf{Component physical predicates.} Measurements use the live physical state. Thresholds shown are implementation defaults; each task stores its parameters and combines the applicable components with any task-bound release, settling, and process requirements, as illustrated by the task examples in Sec.~\ref{app:criteria}.}
  \label{tab:goal-semantics}
  \renewcommand{\arraystretch}{1.10}
  \begin{tabularx}{\linewidth}{@{}>{\raggedright\arraybackslash}p{.21\linewidth}>{\raggedright\arraybackslash}X@{}}
    \toprule
    Predicate & Measurement, rule, and evaluation point\\
    \midrule
    \textbf{Navigation arrival} & \emph{Input:} Measured base pose, target, traversable occupancy. Shortest free-space distance at most 1.5\,m, independently of heading. \emph{Evaluation:} Action boundary; task navigation target.\\
    \addlinespace[4pt]
    \textbf{On a surface or object} & \emph{Input:} Object occupied-geometry bounds; live support footprint and height. Horizontal bounding-box center lies in the support footprint; the lowest occupied point lies at or above the support height minus 0.03\,m. For object support, use that object's live top bounds. \emph{Evaluation:} Current state; named support and support tolerance.\\
    \addlinespace[4pt]
    \textbf{In a surface region} & \emph{Input:} Object occupied points; region axes, extents, and height. All occupied points lie within the horizontal region expanded by 0.03\,m; the same lower-height test applies. \emph{Evaluation:} Current state; named receiving region and support tolerance.\\
    \addlinespace[4pt]
    \textbf{Explicit bounded goal} & \emph{Input:} Object occupied points; a 3D goal box. Every occupied point lies inside all three bounds, expanded by the goal's position tolerance (default 0.02\,m). \emph{Evaluation:} Final state; explicit goal bounds and tolerance.\\
    \addlinespace[4pt]
    \textbf{Inside a cavity} & \emph{Input:} Object center of mass; live cavity frame, footprint, floor, and height. Center of mass is inside the cavity. Side/ceiling allowance is 0.02\,m; floor allowance is 0.05\,m. \emph{Evaluation:} Current state; cavity annotation and moving-parent transform.\\
    \addlinespace[4pt]
    \textbf{Outside a cavity} & \emph{Input:} Occupied points in the live cavity frame. Fraction of sampled occupied points inside is at most 0.05 by default. \emph{Evaluation:} Current state; task's allowed inside fraction.\\
    \addlinespace[4pt]
    \textbf{Open or closed joint} & \emph{Input:} Joint coordinate and semantic endpoints. Normalized openness $r(q)$; default open $r\geq0.75$, closed $r\leq0.15$. Explicit thresholds are applied directly. \emph{Evaluation:} Requested phase; joint target and normalized threshold.\\
    \addlinespace[4pt]
    \textbf{Held or transferred} & \emph{Input:} Per-hand physical contact state, or an enabled assisted binding. Physical possession and handover history as specified in the text; required hands and objects are matched explicitly. \emph{Evaluation:} Per physical sample; effector, grasp state, and overlap requirement.\\
    \addlinespace[4pt]
    \textbf{Route cleared} & \emph{Input:} Updated occupied space; required source and destination. Required locations remain connected; an alternative route is accepted. \emph{Evaluation:} Current layout; task route endpoints.\\
    \addlinespace[4pt]
    \textbf{Explicit pose restored} & \emph{Input:} Measured object pose and requested pose. Position distance at most 0.02\,m and shortest quaternion angular distance at most 0.1\,rad by default. \emph{Evaluation:} Final state; explicit pose goal and tolerances.\\
    \bottomrule
  \end{tabularx}
\end{table}

\begin{samepage}
\paragraph{Geometric interpretation.}
Surface placement uses a spatial convention that permits intervening supports such as a tray or plate.
The \texttt{on\_surface} relation checks the horizontal center of the object's occupied-geometry bounds against the named support footprint and applies the lower-height bound in Table~\ref{tab:goal-semantics}; it neither imposes an independent upper-height bound nor requires direct collider contact.
For \texttt{on\_object}, the receiving footprint and height follow the named object's current geometry, as in the mug-on-board example.
The more specific \texttt{in\_region} relation checks the entire horizontal occupied footprint against the designated region.
Cavity containment is defined by center-of-mass inclusion in the cavity's current local frame, which follows drawer motion.
The storage example's placement and explicit cavity predicates resolve to this same measurement.
An explicitly requested outside condition uses the fraction of occupied points remaining inside.
\par\end{samepage}

\paragraph{Navigation and clearance.}
For base position $\mathbf x$ and target $\mathbf x_g$,
\begin{equation}
    \mathrm{Arrived}(\mathbf x,\mathbf x_g)
    =\mathbf 1\!\left[d_{\mathrm{free}}(\mathbf x,\mathbf x_g)\leq1.5\,\mathrm m\right],
    \label{eq:arrival}
\end{equation}
where $d_{\mathrm{free}}$ is the shortest traversable distance on the task's occupancy representation and is infinite for disconnected locations.

\paragraph{Release and settling.}
Placement release requires that neither hand holds the placed object; another object may remain held.
Handover uses the separate donor-release condition described below.
When stable release is requested, the observation window begins after release and lasts 0.5\,s by default; reacquisition by either hand restarts this window.
Translational motion is considered settled if either the maximum linear speed is at most 0.05\,m/s or the positional span is at most 0.01\,m; the maximum angular speed must also be at most 0.5\,rad/s.
Positional span is the Euclidean norm of the coordinate-wise position range over the window.
These tolerances define stability; task-specific values are stored with the evaluator.
All four placement examples above use the 0.5\,s post-release window; the drawer-opening example requires only its joint outcome.

\paragraph{Physical possession.}
R1 maintains a possession state for each hand--object pair.
Contact-driven possession requires opposed contacts from both fingers on the same object, a closing command with mean finger target at most 0.012\,m, and no reported contact between that object and a non-robot external body.
The contact geometry must have at least 1\,mm inward projection from each finger and 1\,mm separation between the contact centroids along the gripper opening axis.
One qualifying 120\,Hz sample establishes the possession predicate used by the task; any required lift or sustained hold is checked separately.
Assisted execution reads the enabled virtual binding for the specified pair.
Handover requires recipient possession, donor release, and the requested earlier overlap, one physics tick by default.

\paragraph{Joint targets.}
Openness is normalized by the semantic closed and open positions:
\begin{equation}
    r(q)=\operatorname{clip}\!\left(
      \frac{q-q_{\mathrm{closed}}}{q_{\mathrm{open}}-q_{\mathrm{closed}}},0,1\right).
    \label{eq:semantic-joint}
\end{equation}
Generic open and closed conditions use nominal thresholds 0.8 and 0.1 with a 0.05 normalized allowance, yielding $r\geq0.75$ and $r\leq0.15$.
Explicit normalized thresholds are applied directly.

\paragraph{Process monitoring.}
Facts within one event must hold simultaneously; ordered events collect evidence after their predecessors complete.
An invariant records violations over its monitored operation interval.
For joint-operation requirements, the interval starts when measured robot--fixture interaction accompanies normalized joint displacement greater than 0.005 from the waiting-state origin, and ends at the requested joint target.
Pickup, staging, and delivery outside this interval are excluded.
In the operate-while-carrying example, left-hand possession is required during the door-opening interval, while the final receiving-region conditions require the cube to be released and settled.
Physical interaction is contact with the moving rigid assembly at a reported separation of at most 1\,mm; assisted execution uses an active binding to that assembly.
The clear-before-motion predicate checks the designated obstruction against the live fixture-frame sweep at the first such interaction-driven motion and retains that result.
These monitors follow measured effects and observed predecessor completion, independently of the reference program counter.

\subsection{Dependency-Aware Progress}
\label{app:progress}

Let $c_i[k]\in\{0,1\}$ indicate whether semantic endpoint $i$ has earned credit by evaluation sample $k$.
Its scored prerequisites $\operatorname{pred}(i)$ are obtained by projecting operation-graph reachability onto scored endpoints and taking the transitive reduction of the resulting relation.
Contracting technical bridges preserves the partial order among scored semantic endpoints; the bridges receive zero weight.
For predicate result $e_i[k]$, evaluated using evidence collected after activation,
\begin{equation}
    c_i[k]=c_i[k-1]\lor
    \left(e_i[k]\land\bigwedge_{j\in\operatorname{pred}(i)}c_j[k-1]\right).
    \label{eq:endpoint-completion}
\end{equation}
Initially $c_i[0]=0$, with predecessor-free endpoints active from the start.
Independent ready endpoints are monitored in parallel.
The previous-sample dependency check prevents a single observation from proving successive operations, and completed endpoints retain their credit.

For positive-weight endpoints $V_+$,
\begin{equation}
    P=\frac{\sum_{i\in V_+}w_i c_i}{\sum_{i\in V_+}w_i}.
    \label{eq:supp-progress}
\end{equation}
Weights are fixed at compilation and shared across systems; zero-weight technical transitions are excluded.
Complete-task success is evaluated separately, leaving $P$ unchanged.

\paragraph{Weight convention.}
Table~\ref{tab:progress-weights} lists the operation weights and increments.
An operation receives its base weight, the largest applicable geometric increment, and the applicable coordination and loaded-navigation increments.
An endpoint's weight is the sum of its associated operation weights, plus 0.20 for participation in an explicit synchronization group.
Acquisition credit for an object required to remain held at completion has a minimum weight of 1.2.
Termination and score aggregation follow Sec.~\ref{app:termination}.

\begin{table}[tp]
  \centering\small
  \caption{\textbf{Semantic progress weights.} Values are fixed at task compilation. The geometric contribution is the largest applicable increment; coordination and loaded-navigation contributions are additive.}
  \label{tab:progress-weights}
  \renewcommand{\arraystretch}{1.10}
  \begin{tabularx}{\linewidth}{@{}>{\raggedright\arraybackslash}p{.23\linewidth}>{\raggedright\arraybackslash}X r@{}}
    \toprule
    Component & Applicable case & Weight\\
    \midrule
    Base operation & Navigation & 1.35\\
     & Acquisition & 1.00\\
     & Placement & 1.00\\
     & Fixture actuation & 1.20\\
     & Hand transfer & 1.60\\
    \addlinespace[3pt]
    Geometry (maximum) & Relative or regional target & +0.10\\
     & Container contents or object-supported interaction & +0.15\\
     & Storage interaction & +0.25\\
     & Floor-level large container & +0.30\\
     & Fixture actuation while holding a payload & +0.20\\
     & Ordinary surface interaction or empty-hand fixture actuation & 0\\
    \addlinespace[3pt]
    Coordination & Independent two-hand handling & +0.10\\
     & Cooperative or synchronized handling & +0.30\\
    \addlinespace[3pt]
    Loaded navigation & Navigation while carrying an object & +0.25\\
    \addlinespace[3pt]
    Endpoint synchronization & Participation in an explicit synchronization group; add to the sum of associated operation weights & +0.20\\
    \addlinespace[3pt]
    Held-object acquisition & Object required to remain held at task completion & Minimum 1.20\\
    \bottomrule
  \end{tabularx}
\end{table}

\section{Experience Records and Reuse}
\label{app:learning}

\subsection{Recorded Failure and Executed Program Revision}
\label{app:program-revision}

Figure~\ref{fig:execution-experience}(b) uses two independent RESET trials of \texttt{clear\_and\_place\_4805082873374871}.
An operator supplied the first attempt's retained adjustment record to the second attempt; the agent then selected and executed the revised actions from fresh measurements.
The reused record identified a grasp that passed a short lift but retained contact near the pad tips and slipped during carry preparation.
In the second trial, the agent detected the same shallow-contact tendency, supported the board on the table, opened its fingers, and measured a board yaw of 0.15724\,rad.
It wrote and executed the short helper below to generate a yaw-aligned horizontal wrist target through the existing native pose planner.

\begin{figure}[t]
  \centering
  \begin{minipage}{\linewidth}
  \footnotesize
\begin{verbatim}
from episode import call
import math
a = .15724 / 2
q = [math.cos(a)/math.sqrt(2),
     math.cos(a)/math.sqrt(2),
     math.sin(a)/math.sqrt(2),
     math.sin(a)/math.sqrt(2)]
call('plan_pose', goals={'right': {
    'position_m': [.241, -8.87, .7565],
    'orientation_wxyz': q}},
    allow_base_motion=False,
    allow_shared_torso=True)
\end{verbatim}
  \end{minipage}
  \caption{\textbf{Executed reseating helper.} The recorded \texttt{board\_reseat\_v1.py} call, with line wrapping changed for display. The pose uses the board yaw measured at step 6515; the native planner reaches the target at step 6932. Closure, trial lift, and transport follow as separately recorded actions.}
  \label{fig:revision-helper}
\end{figure}

After reseating, closure produced opposed finger contacts spanning $[-34.8,17.3]$\,mm along the wrist-frame pad-depth axis, compared with $[10,17]$\,mm near the tips in the earlier grasp.
During the subsequent approximately 42\,mm lift, the measured board-relative depth remained near 90.2\,mm; acquisition exit and native transport preserved the grasp.
The complete second trial placed the board and mug in their required regions and passed the original evaluator after release and withdrawal.
The two records contain 158.43\,s and 450.47\,s of simulated execution, respectively.

\subsection{Complete Carrying-and-Articulation Record}
\label{app:carrying-case}

The synchronized runtime views and contact geometry in Fig.~\ref{fig:mechanisms}(b--c) come from one continuous InteriorGS \texttt{0180\_840102} collection trial.
R1 Pro acquires a toy bus, carries it while using the free hand to close a microwave, then returns to place and release the bus.
The two commanded hinge increments are $-0.70$ and $-0.88782984$\,rad; the final measured hinge angle is $0.00324077$\,rad ($0.1857^\circ$).
The original evaluator reports task success after withdrawal at step 8008 (266.93\,s), with no virtual grasp bindings.
Ego, third-person, and both wrist-camera videos, action records, measured states, and three completed contact segments are retained.

\begin{figure}[t]
  \centering
  \includegraphics[width=\textwidth]{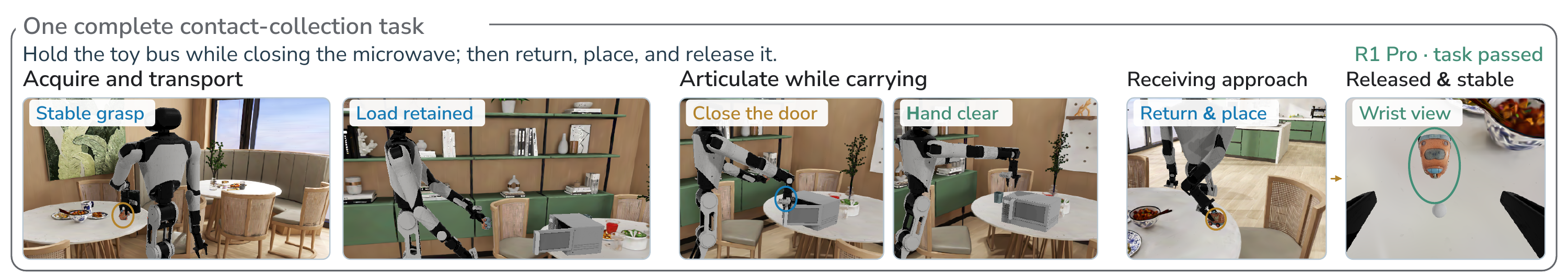}
  \caption{\textbf{One continuous contact-collection trial.} Acquisition and carrying of a toy bus, free-hand microwave closure and withdrawal, and final placement and release. The original task evaluator reports success after the final withdrawal. All panels use measured states from the same trial, without virtual grasp bindings.}
  \label{fig:carrying-record}
\end{figure}

\subsection{Recording and Temporal Alignment}
\label{app:visual-examples}

Immersive teleoperation supplies wrist and hand commands; browser-supported control provides observation, robot selection, and navigation interaction.
An operator can record a whole task or a selected segment.

Records retain observations, executed commands, measured robot states and outcomes, timestamps, operation boundaries, and control source.
Commands and measured outcomes occupy separate fields, so tracking errors and unsuccessful contact remain visible.
Assistance labels identify virtual bindings; agentic records additionally retain goals, tool arguments, and returned feedback.

For demonstration export, each action is paired with the preceding robot state and camera frames captured no later than that state's timestamp.

\subsection{Source Selection and Outcome Labels}

Export uses an explicit source list, optionally with start and end simulation times (Table~\ref{tab:export-selection}).
Whole-recording labels come from task results or teleoperation annotations.
Each selected interval has a separately specified outcome or is labeled as unknown; the outcome of the full recording is retained independently.

\begin{table}[htbp]
  \centering\small
  \caption{\textbf{Source selection, outcome labels, and assistance filtering.}}
  \label{tab:export-selection}
  \renewcommand{\arraystretch}{1.12}
  \begin{tabularx}{\linewidth}{@{}L{.31\linewidth}Y@{}}
    \toprule
    Selected source & Export behavior\\
    \midrule
    Complete episode & Retain outcome, control source, and assistance for each selected record, including failed episodes.\\
    Explicit interval & Export its transitions; retain the full-recording outcome separately. Segment success is supplied or unknown.\\
    Physical-only sources & Retain only explicitly unassisted sources; exclude assisted and unknown-assistance sources.\\
    Mixed sources & Preserve source and assistance labels across agent, human, and reference records.\\
    Incomplete recording & Preserve incomplete status; require usable action/state pairs and requested camera frames.\\
    \bottomrule
  \end{tabularx}
\end{table}

By default, export retains the assistance categories represented in the source list; enabling physical-only filtering restricts export to explicitly unassisted sources.

\paragraph{Image credits.}
Figure~\ref{fig:overview}: Vision Pro photograph by \href{https://commons.wikimedia.org/wiki/File:Vision_Pro_-_Facing_left.jpg}{Hay Kranen} (\href{https://creativecommons.org/licenses/by/4.0/}{CC BY 4.0}); glove photograph by MANUS.

\section{Evaluation Protocol and Baseline Interfaces}
\label{app:protocol}

\subsection{Suite Inputs and System Interfaces}

A suite manifest identifies compiled instances by task ID, relative package location, and reset seed.
Generation seeds, reset seeds, and policy randomness have separate roles; loading a compiled instance restores its fixed layout.
In ASPIRE's suite adapter, the argument named seed indexes the ordered instance list; the separate package reset seed controls physical initialization.
ASPIRE's skill library and RAX's exploration artifacts are developed on separate instances outside the fixed 220-instance suite and frozen before formal evaluation.
RAX uses development packages from the same task families.

\paragraph{Shared evaluation settings.}
\begin{description}
  \setlength{\itemsep}{3pt}
  \item[Embodiment and suite.] R1 Pro; 220 instances, with ten per family; all retained instances completed by the reference Oracle with manipulation assistance.
  \item[Repetitions.] Three valid trials per system--task pair, replacing invalid runs until three valid results are obtained; retain one trial by success first and semantic progress second. Both metrics use that trial.
  \item[Physical runtime.] 120\,Hz physics, 30\,Hz control; physics pauses between actions, including during model inference.
  \item[Initialization.] Each trial loads the same compiled task state and its package reset seed; the scene layout remains fixed across policy seeds.
  \item[Development inputs.] ASPIRE's skill library and RAX's exploration artifacts come from separate development instances outside the fixed 220-instance suite and are frozen before formal evaluation; RAX uses same-family packages.
  \item[Evaluation memory.] Episode-produced histories, memory, and program revisions are reset between trials; the frozen development inputs are restored at each trial's start.
  \item[Manipulation assistance.] Virtual grasp bindings and fixture attachments are disabled for all four baseline systems; grasping and articulation depend on physical contact.
\end{description}

$B$ denotes the task-scaled allocation in Eq.~\ref{eq:execution-budget}.

Agent-visible inputs are distinguished from execution-tool geometry below.

\paragraph{CaP-X.}
\emph{Inputs and execution.} Instruction, head and wrist RGB-D, calibration, and base/joint/hand state through program APIs. Planning combines authored static geometry with observed depth geometry. Generated programs invoke segmentation, grasp proposals, navigation, wrist/joint motion, torso, and gripper tools. Subsequent reasoning receives tool results, observations, and program output.

\emph{Model and perception.} GPT-6-Astra for program generation and visual differencing; temperature 1; at most 20,480 output tokens per response. Molmo2-8B~\citep{molmo2}, SAM 3~\citep{sam3}, and Contact-GraspNet~\citep{contactgraspnet}; run-local programs and execution feedback.

\emph{Memory and limits.} The adapter maintains run-local programs and execution feedback. $\lceil30B\rceil$ control actions; ten program-revision rounds.

\paragraph{ASPIRE.}
\emph{Inputs and execution.} Instruction, head and wrist RGB-D in an odometry frame, and robot state. Planning uses authored static geometry and public depth observations. Programs invoke perception and motion providers, receiving structured plans, execution results, and fresh pre/post-motion keyframes.

\emph{Model and perception.} GPT-6-Astra; \texttt{xhigh} reasoning effort. SAM 3~\citep{sam3} and Contact-GraspNet~\citep{contactgraspnet}; program development reuses a frozen skill library.

\emph{Memory and limits.} The skill library is developed on instances outside the fixed 220-instance suite and frozen before evaluation; task-conditioned program revisions remain local to each trial. Package action horizon, bounded by $\lfloor30T_{\max}\rfloor$; no additional attempt wall-time cap in the launcher.

\paragraph{OWMM-Agent.}
\emph{Inputs and execution.} Decision model: instruction, current RGB image, eight tour images, and textual history. Execution adapter: depth, calibration, robot state, and configured collision geometry. One structured action per turn: tour-frame navigation, point navigation, pick, or place. Depth grounds image boxes; the next turn receives the new view and model-produced summary.

\emph{Model and perception.} Released 8B checkpoint (\texttt{hhyhrhy/OWMM-Agent-Model}); temperature 0, top-$p$ 0.5; 1,024 output tokens. Eight tour images, current RGB and per-episode textual history; policy seeds 1, 2, 3.

\emph{Memory and limits.} Tour frames are loaded and shuffled with the policy seed. Textual history starts empty each episode and is updated using the model's summary. At most 20 model turns; $B$ execution-wall seconds, excluding model queries.

\paragraph{RAX.}
\emph{Inputs and execution.} Instruction, head and wrist RGB-D with calibration, base-relative observations, and 61-D proprioception. The native free-space arm tool uses a self-collision-only model. Upstream BEHAVIOR tools use planned motions, local end-effector control, and 23-D commands at 30\,Hz. The harness returns observations and command results.

\emph{Evaluation feedback.} A binary task-completion query is restricted to development episodes. Formal episodes expose robot tools; terminal action replies report episode termination and success. The runner reads the final evaluation after execution. Numerical semantic progress and detailed unsatisfied predicates are available only in evaluator records.

\emph{Model and perception.} GPT-6-Astra; remaining decoding options inherited from its agent harness. Exploration artifacts and agent-written skills passed to formal execution.

\emph{Memory and limits.} Separate same-family development packages, outside the 220-instance suite, supply exploration artifacts frozen before formal evaluation. Scored-trial memory remains local to each trial. $\lceil30B\rceil$ control actions; no episode wall-time cap.

\subsection{Termination, Budget Clocks, and Scoring}
\label{app:termination}

The standard environment evaluates physical samples during each action and determines termination at the action boundary.
Automatic success and irrecoverable-failure termination are enabled by default; final scoring uses the latest measured outcome and retained process and endpoint history.
Contact collection uses the deferred-success procedure in Sec.~\ref{app:collection-completion}.

The final evaluator state and accumulated endpoint credit are frozen when the episode ends.
Runner-side stop reasons are retained separately from the environment's terminal reason.

\begin{description}
  \setlength{\itemsep}{3pt}
  \item[Automatic success.] At the action boundary, current success satisfies the configured consecutive-success count and success termination is enabled. Default count: one action. Valid success; accumulated semantic progress is retained.
  \item[Agent / runner finish.] Return an already finalized report unchanged. If the episode is still running without qualifying success, close it as a policy stop. In explicitly deferred mode, current stable success can be finalized. Success only if finalized from current evaluator success; otherwise valid failure with accumulated progress.
  \item[Irrecoverable task violation.] An explicitly monitored irreversible condition fails and failure termination is enabled. This takes precedence over success. Valid failure with accumulated semantic progress.
  \item[Simulation-time or action limit.] At the action boundary, apply the enabled limit. Qualifying current success takes precedence over a limit reached on that same boundary; otherwise truncate. Success if already achieved under that precedence; otherwise valid failure with accumulated progress.
  \item[Invalid physical action.] A decoded action violates the configured action space or joint limits before execution. Valid failure with previously accumulated semantic progress.
  \item[Explicit disqualification.] A configured safety-violation terminal category is invoked. Ordinary incidental contact alone does not create this category. Disqualified; success is false and progress is zero.
  \item[Simulation failure.] Physical runtime, observation, or evaluator processing fails. Invalid run; success and progress are recorded as unavailable.
  \item[Model / transport failure.] The runner records the service or infrastructure failure and cleans up the worker. The runner's error classification determines run validity, independently of any cleanup-generated stop report.
\end{description}

\paragraph{Physical limits.}
Each package specifies a simulation-time limit and may also specify an action-count limit. The execution profile determines which limits are enabled.
These limits count physical execution at the rates in Sec.~\ref{app:runtime}, excluding model waiting.
The consecutive-success count is separate from the physical settling window of a placement predicate.

\paragraph{Runner budgets.}
The available task-scaled allocation is
\begin{equation}
    B=2\bigl(9.6L+65M\bigr)+300+300\max(A,1),
    \label{eq:execution-budget}
\end{equation}
where $B$ is in seconds, $L$ is required reference route length in meters, $M$ counts acquisition, placement, fixture-actuation, and hand-transfer operations, and $A$ counts acquisitions.
$B$ is rounded to 0.1\,s.

CaP-X and RAX convert $B$ to $\lceil30B\rceil$ control actions at 30\,Hz, counting simulated execution; model waiting and loaded-GPU wall time fall outside this budget.
CaP-X separately permits ten program-revision rounds and uses a wall-clock watchdog only for stalled execution.
OWMM applies $B$ to execution wall time with measured model-query time excluded, and allows at most 20 model turns.
ASPIRE uses the smaller of an available package action limit and $\lfloor30T_{\max}\rfloor$, where $T_{\max}$ is the package simulation-time limit.
Its attempt launcher leaves per-attempt wall time uncapped; outer service timeouts remain infrastructure safeguards.

\paragraph{Aggregation.}
For each system--task pair, three valid trials are obtained and one result is retained according to lexicographic order: successful trials rank above unsuccessful trials, and semantic progress breaks ties among trials with the same success status.
Success and semantic progress are aggregated over one retained result per task.
The denominator is 220 task instances overall and ten instances within each family.
Overall success therefore equals the mean of the 22 family rates; group summaries are weighted by instance count.
Task-budget exhaustion and unsatisfied goals are scored as unsuccessful trials.
Interruptions that prevent a valid evaluator result are recorded separately from scored task outcomes and do not count toward the three trials.
Invalid runs are replaced until three valid results are available for every system--task pair.
CaP-X's launcher resumes unfinished trials after relay outages and permits one retry for designated service faults, preserving completed outcomes.
This recovery rule applies exclusively to infrastructure faults and is separate from policy repetitions; task failures and budget exhaustion retain their scored outcomes.

\paragraph{Evaluation-time memory.}
Each evaluation trial starts from its initial agent context and the same compiled physical state.
Histories, new memory, and program revisions produced during a scored trial remain local to that trial and are reset between trials.
Within-trial reasoning and feedback remain available to the system throughout that episode.
The frozen development-stage skills or exploration artifacts are restored as initial inputs for each scored trial.

\paragraph{Assistance and action interfaces.}
All four baseline integrations use public observations and contact-driven execution, with virtual grasp and fixture attachments disabled.
OWMM-Agent's prompted vocabulary comprises tour-frame navigation, point navigation, pick, and place; CaP-X, ASPIRE, and RAX expose lower-level motion calls that permit agent-written interaction sequences.

\subsection{Recorded Baseline Executions}
\label{app:baseline-case}

Table~\ref{tab:baseline-case} follows a recorded RAX formal-evaluation episode of \texttt{storage\_retrieval\_7287676246192763}, a member of the fixed 220-instance suite, in InteriorGS scene \texttt{0675\_841498}.
The instruction requires retrieving a blue-and-white box from an already open nightstand compartment, carrying it to the dining table beside the flowers, and placing it securely.
The November 13 record contains public head and wrist RGB-D observations, depth-localization calls, robot-state feedback, and executed motion commands.

\begin{table}[htbp]
  \centering\small
  \caption{\textbf{A recorded RAX storage-retrieval episode.}}
  \label{tab:baseline-case}
  \renewcommand{\arraystretch}{1.12}
  \begin{tabularx}{\linewidth}{@{}L{.21\linewidth}Y@{}}
    \toprule
    Control step & Recorded action and feedback\\
    \midrule
    3334 & Arm-only approach preview fails IK; a higher approach using the torso returns a plan.\\
    3473--3947 & Wrist RGB-D refines the target; the left hand closes, lifts, and retracts the box into a carrying pose.\\
    3947--6348 & Base translations and turns carry the box to the dining table while retaining the gripper command.\\
    6348--6503 & Head and wrist depth guide placement; the hand lowers the box in successive planned motions.\\
    6525 & Finger opening completes secure placement. The evaluator terminates with success and progress 1.0.\\
    \bottomrule
  \end{tabularx}
\end{table}

\paragraph{Grip loss during storage deposit.}
In \texttt{storage\_deposit\_3583950262585049} (InteriorGS \texttt{0173\_840081}), RAX must carry a blue-and-brown pyramid puzzle from a table to an open bookcase compartment.
The evaluator records acquisition, and the wrist image at step 4394 shows the puzzle between the fingers.
A lateral departure stalls at step 4628; backing away succeeds, but the following turn ends at step 4897 with empty, fully closed fingers despite the unchanged closing command.
The agent notices the loss and searches for the object.
A later recovery rotation triggers the wheel-support failure stop at step 5609 (186.97\,s).
Placement and cavity containment remain false, while acquisition retains progress of 0.465.
The physical termination precedes the 30,609-step action budget.
At the last recorded sample, base tilt is 0.97\,rad and two wheels have ground clearances of 0.45 and 0.10\,m.

\paragraph{Recovery motion and physical stability.}
In \texttt{spatial\_arrangement\_2184991977419149} (InteriorGS \texttt{0195\_840124}), the task requires moving a pale green egg to a specified side of a dining table.
Acquisition is credited, but placement and the destination-workspace condition are not.
During recovery, successive end-effector requests finish with position errors of 0.516 and 0.215\,m at steps 16951 and 17071.
A subsequent pose request brings the end effector within 0.0059\,m of its target without completing the task.
At step 17696, the free-space planner returns a self-collision-free plan; execution stops at step 17726 when the robot loses wheel support.
The last recorded base tilt is 1.04\,rad and height drift is 0.29\,m.
The task ends at 590.87\,s with progress 0.5, before its 27,801-step budget.
In both episodes, physics advances through issued actions and measured wheel-support loss triggers termination; the evaluator records valid task failures.
Grip loss and pose errors precede this stopping event.
The collision model used by RAX's free-space tool is limited to robot self-collision (Sec.~\ref{app:protocol}).

\end{document}